\documentclass[lettersize,journal]{IEEEtran}
\usepackage{amsmath,amsfonts}
\usepackage{algorithmic}
\usepackage{algorithm}
\usepackage{array}
\usepackage[caption=false,font=normalsize,labelfont=sf,textfont=sf]{subfig}
\usepackage{textcomp}
\usepackage{stfloats}
\usepackage{url}
\usepackage{hyperref}       
\usepackage{url}            
\usepackage{booktabs}       
\usepackage{amsfonts}       
\usepackage{nicefrac}       
\usepackage{microtype}      

\usepackage{float}
\RequirePackage[table]{xcolor} 
\usepackage{makecell}       
\usepackage[utf8]{inputenc} 
\usepackage[T1]{fontenc}    
\usepackage{hyperref}       
\usepackage{url}            
\usepackage{booktabs}       
\usepackage{amsfonts}       
\usepackage{nicefrac}       
\usepackage{microtype}      
\usepackage{xcolor}         
\usepackage{booktabs}
\usepackage{multirow}
\usepackage{makecell}
\usepackage{enumitem}
\usepackage{bm}
\usepackage{amssymb}
\usepackage{graphicx}
\usepackage{enumitem}
\usepackage{amsmath}
\usepackage{wrapfig}
\usepackage{xcolor}
\usepackage{enumitem}
\usepackage{caption}
\usepackage{verbatim}
\usepackage{graphicx}
\usepackage{cite}
\definecolor{mycustompurple}{RGB}{154, 36, 79} 
\usepackage[utf8]{inputenc}

\usepackage{hyperref}

\hypersetup{
	colorlinks=true,            
	linkcolor=red,             
	citecolor=green,            
	filecolor=mycustompurple,  
	urlcolor=magenta           
}
\begin{document}

\title{SplatWeaver: Learning to Allocate Gaussian Primitives for Generalizable Novel View Synthesis}

\author{Yecong Wan, Fan Li, Mingwen Shao~\IEEEmembership{Member,~IEEE}, and Wangmeng Zuo~\IEEEmembership{Senior Member,~IEEE}
\thanks{Yecong Wan and Wangmeng Zuo are with the Faculty of Computing, Harbin Institute of Technology, Harbin, 150001, China. Yecong Wan is also with the Zhengzhou Advanced Research Institute of Harbin Institute of Technology, Zhengzhou, 450000, China.}
\thanks{Fan Li is with the Huawei Noah’s Ark Lab, Shenzhen, 518100, China.}
\thanks{Mingwen Shao is with the Artificial Intelligence Research Institute, Shenzhen University of Advanced Technology, Shenzhen, 518107, China.}

}

\markboth{Journal of \LaTeX\ Class Files,~Vol.~14, No.~8, August~2021}%
{Shell \MakeLowercase{\textit{et al.}}: A Sample Article Using IEEEtran.cls for IEEE Journals}


\maketitle

\begin{abstract}

Generalizable novel view synthesis aims to render unseen views from uncalibrated input images without requiring per-scene optimization. Recent feed-forward approaches based on 3D Gaussian Splatting have achieved promising efficiency and rendering quality. However, most of them assign a fixed number of Gaussians to each pixel or voxel, ignoring the spatially varying complexity of real-world scenes. Such uniform allocation often wastes Gaussian primitives in smooth regions while providing insufficient capacity for fine structures, complex geometry, and high-frequency details. This motivates us to predict region-dependent primitive cardinalities rather than impose a fixed primitive budget everywhere, enabling a more expressive yet compact 3D scene representation.
Therefore, we propose SplatWeaver, a generalizable novel view synthesis framework that is able to dynamically allocate Gaussian primitives over different regions in a feed-forward manner. Specifically, SplatWeaver introduces cardinality Gaussian experts and a pixel-level routing scheme, wherein each expert specializes in producing a specific number of primitives from 0 to M, and the routing scheme coordinates these experts to adaptively determine how many Gaussian primitives should be allocated to each spatial location.
Moreover, SplatWeaver incorporates a high-frequency prior with attendant guidance module and routing regularization to stabilize expert selection and promote complexity-aware allocation. By leveraging high-frequency structural cues, the routing process is encouraged to assign more Gaussian primitives to fine structures, complex geometry, and textured regions, while suppressing redundant primitives in smooth areas. This results in a ``\emph{dense where complex, sparse where smooth}'' allocation behavior. Extensive experiments across diverse scenarios show that SplatWeaver consistently outperforms state-of-the-art methods, delivering more faithful novel-view renderings with fewer Gaussian primitives. Project Page: \url{https://yecongwan.github.io/SplatWeaver/}.
%
%
\end{abstract}


\begin{figure}[h]
	\begin{center}
		\includegraphics[width=\linewidth]{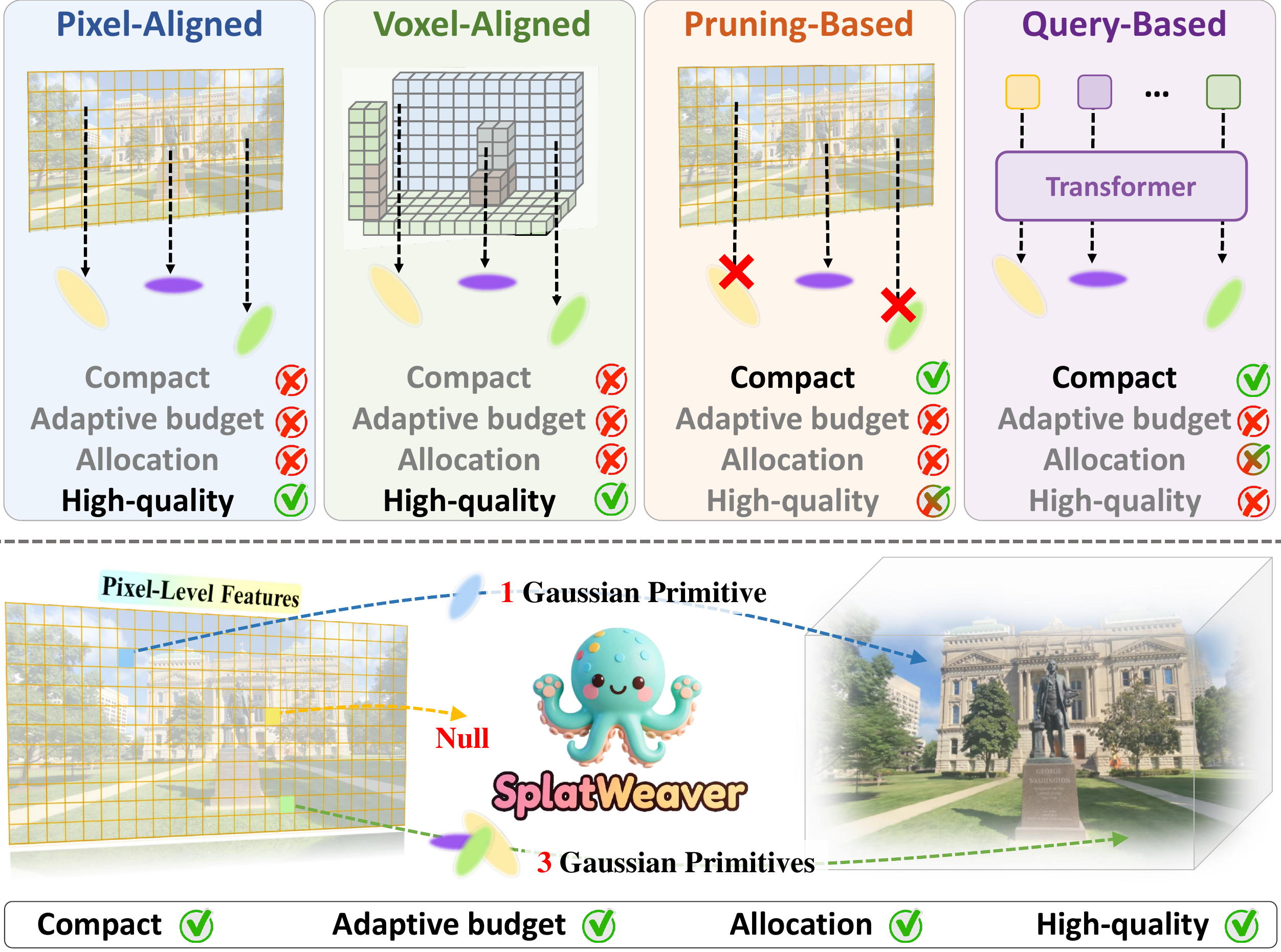}
	\end{center}
	\vspace{-6pt}
	\caption{ \textbf{Comparison of paradigms for generalizable novel view synthesis}. In contrast to prior methods that struggle with redundant primitives, fixed budgets, or rigid allocation,
    SplatWeaver adaptively allocates a dynamic number of Gaussian primitives according to scene complexity, enabling a more principled and flexible distribution of scene representations. }
	\label{figure_com}
	\vspace{-6pt}
\end{figure}

\section{Introduction}
The pursuit of photorealistic 3D scene creation has evolved from handcrafted pipelines to fully differentiable models that learn directly from raw image observations. This evolution has been catalyzed by the emergence of powerful neural representations such as Neural Radiance Fields (NeRF) \cite{mildenhall2021nerf} and 3D Gaussian Splatting (3DGS) \cite{kerbl20233d}, which have dramatically pushed the boundaries of novel view synthesis. The success of these breakthroughs and their variants \cite{chen2022tensorf, fridovich2022plenoxels, garbin2021fastnerf, muller2022instant,yu2024mip, cheng2024gaussianpro, wu20244d, lu2024scaffold} has sparked a surge of research for generalizable novel view synthesis \cite{charatan2024pixelsplat,chen2024mvsplat,ziwen2025long,jiang2025anysplat,xu2023wavenerf}, seeking to eliminate costly scene-specific optimization.

Earlier paradigms aimed to directly reconstruct scene geometry and appearance from pre-calibrated viewpoints, spanning from sparse dual-view configurations \cite{charatan2024pixelsplat,chen2024mvsplat,xu2025depthsplat,zhang2024gs,min2024epipolar,tang2024hisplat} to dense sequences comprising hundreds of views \cite{ziwen2025long,wang2025zpressor}, demonstrating impressive novel view synthesis performance. However, the assumption of known camera poses is often infeasible in unconstrained or "in-the-wild" scenarios, significantly hindering the practical utility and robustness of these approaches.
To this end, recent research \cite{jiang2025anysplat,ye2026yonosplat,ye2024no,zhang2025flare,hong2024pf3plat,smart2024splatt3r} has sought to construct more robust feed-forward reconstruction models that jointly estimate camera poses and 3D representations directly from uncalibrated observations, thereby enabling more generalized novel view synthesis in unconstrained environments.

Despite these advances, the majority of existing methods rely on either pixel-aligned \cite{charatan2024pixelsplat,ziwen2025long,chen2024mvsplat,xu2025depthsplat} or voxel-aligned \cite{miao2025evolsplat,wang2025volsplat,jiang2025anysplat,li2026tokensplat,liu2025worldmirror} Gaussian prediction schemes. Such uniform paradigms lack the adaptive pruning and densification strategies inherent in vanilla 3DGS \cite{kerbl20233d}, preventing dynamic adjustment of Gaussian distribution across regions of varying complexity. Consequently, this leads to structural redundancy in smooth areas, such as flat walls, while causing under-fitting in regions with intricate textures and complex geometry.
To mitigate the excessive growth of Gaussians caused by high-resolution dense views, several methods have explored opacity-based pruning \cite{ziwen2025long,ye2026yonosplat,zhang2024gaussian,park2025ecosplat} or early truncation \cite{moreau2025off,singhalgaussiantrim3r}. However, they still fail to adaptively reallocate Gaussian primitives across varying scene complexity. Although recent methods such as C3G \cite{an2025c3g} and TokenGS \cite{tokengs2026} introduce token querying mechanisms to predict Gaussian distributions, their reliance on a predefined number of tokens inherently limits their adaptive scalability across diverse scenes and varying levels of view coverage.
Nevertheless, while these approaches can partially control the total number of Gaussians, they lack the flexibility to dynamically allocate primitives with adaptive budgets, leading to sub-optimal primitive distribution and compromised rendering quality.

\begin{figure*}[!t]
	\centering
	\includegraphics[width=\textwidth]{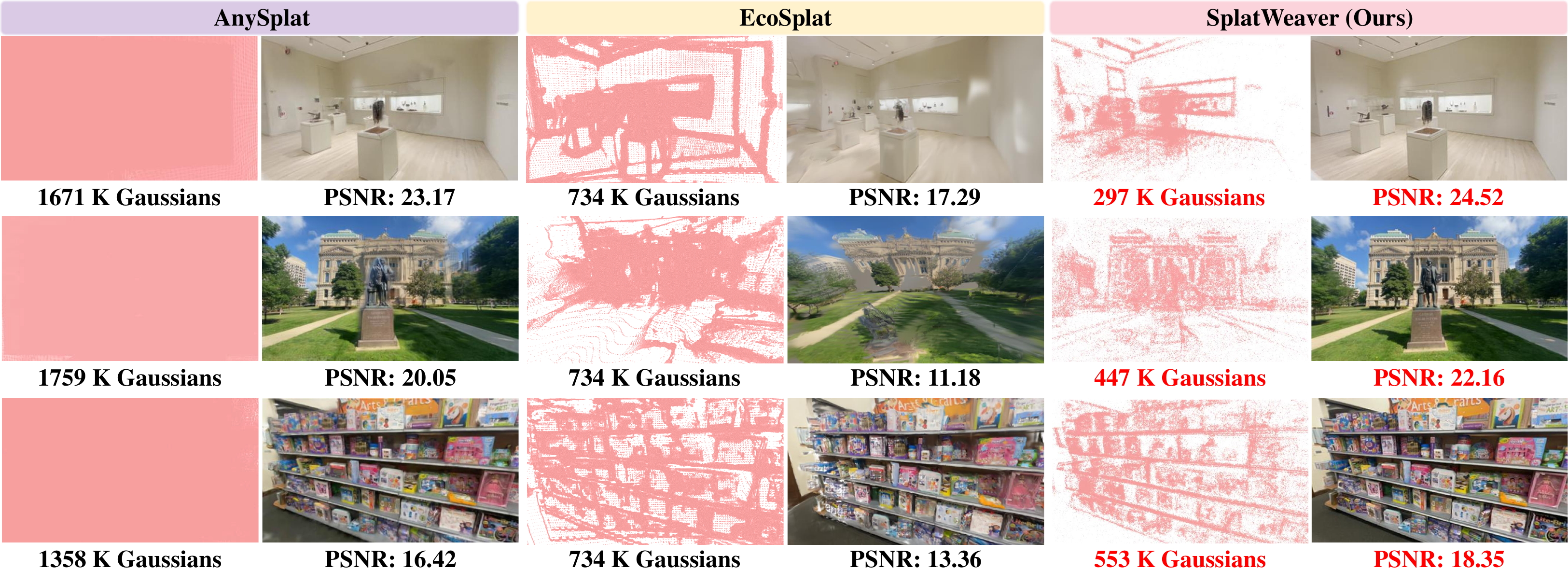}
	\vspace{-12pt}
	\caption{Comparison of predicted Gaussian distributions and novel view synthesis performance. SplatWeaver dynamically distributes Gaussians across different spatial regions in accordance with scene complexity. By concentrating primitives in intricate areas while maintaining sparsity in smooth regions, it achieves higher-quality rendering with a more compact representation.}
	\label{figure_vis}
	\vspace{-7pt}
\end{figure*}

\begin{figure}[h]
	\begin{center}
		\includegraphics[width=.9\linewidth]{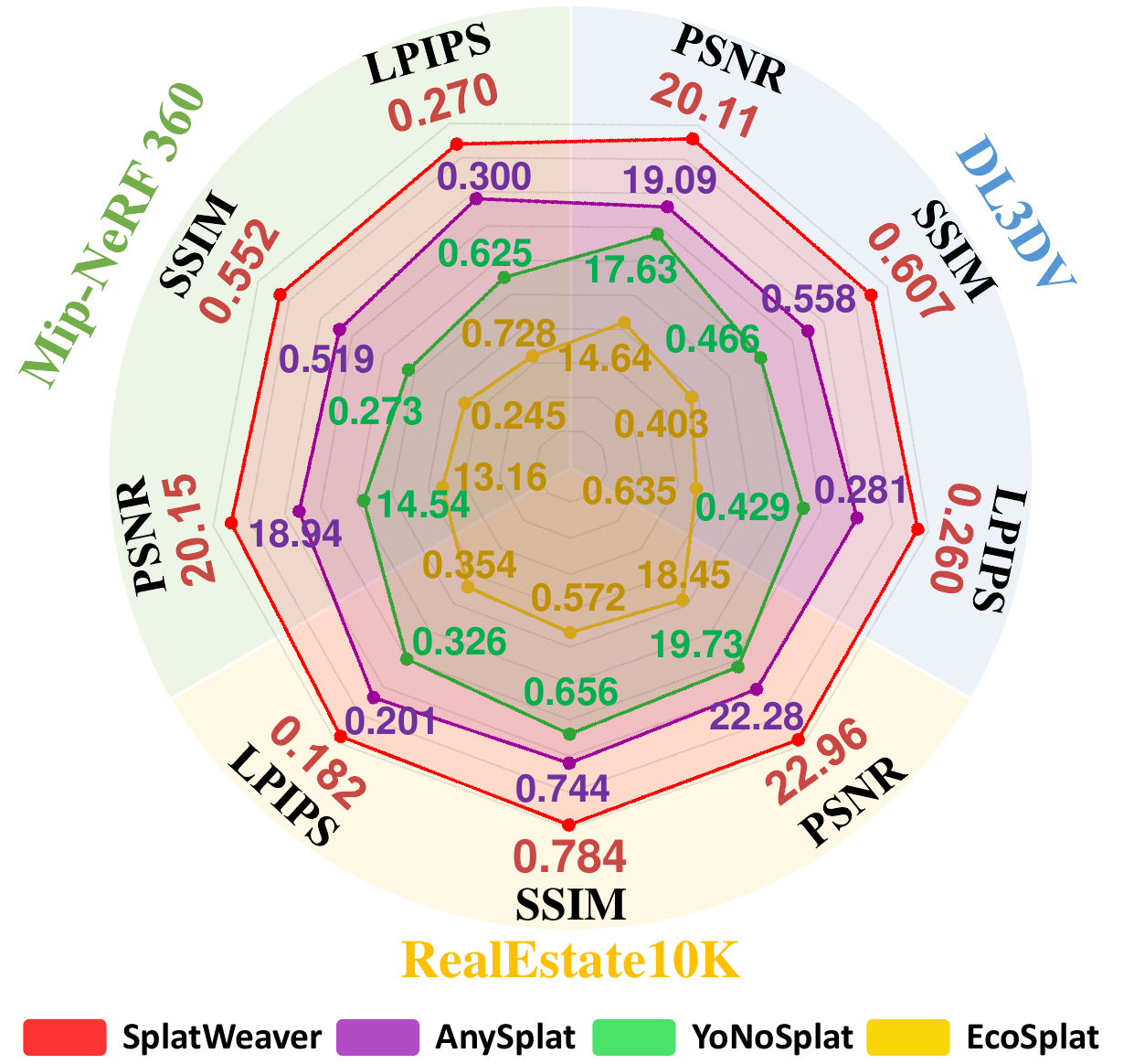}
	\end{center}
	\vspace{-7pt}
	\caption{ SplatWeaver achieves consistent state-of-the-art performance across three benchmarks in pose-free generalizable novel view synthesis.}
	\label{figure_radar}
	\vspace{-6pt}
\end{figure}

To address the aforementioned limitations, we propose \textbf{SplatWeaver}, 
an innovative framework that adaptively allocates Gaussian primitives based on scene complexity in a feed-forward manner, enabling more efficient and high-fidelity generalizable novel view synthesis (Fig. \ref{figure_com}).
Specifically, we introduce the concept of cardinality Gaussian experts, wherein each expert is specialized in predicting a specific number of Gaussian primitives (ranging from 0 to $M$). Complemented by a pixel-level routing scheme, this framework enables the flexible allocation of Gaussian primitives across the scene. Instead of directly regressing complete Gaussian parameters, each expert predicts a set of hidden Gaussians comprising spatial positions and associated latent features. These are subsequently aggregated with spatial neighborhood context to derive the final parameters, yielding more coherent and precise primitive attributes.
Furthermore, to stabilize expert routing, we leverage a high-frequency prior and introduce a frequency prior guidance module alongside a routing regularization term, facilitating a more complexity-aware and structurally sound allocation.
Extensive experiments across a diverse range of scenarios substantiate that SplatWeaver can allocate Gaussian primitives with superior flexibility and efficacy. Our approach yields more coherent and faithful renderings, consistently outperforming alternatives both quantitatively and qualitatively (Fig. \ref{figure_vis} and Fig. \ref{figure_radar}). 
Furthermore, SplatWeaver also exhibits an emergent allocation capability: it can automatically adjust the Gaussian budget according to view coverage and scene complexity, revealing remarkable versatility and practicality.

In conclusion, the main contributions are summarized as follows:

\begin{itemize}[left=4pt, nosep]
	\item We propose a novel framework, termed SplatWeaver, which enables adaptive allocation of Gaussian primitives according to scene complexity in a feed-forward manner, significantly advancing both the efficiency and rendering quality of generalizable novel view synthesis.
	
	\item We introduce the concept of cardinality Gaussian experts and employ a dedicated pixel-level routing mechanism to enable flexible and adaptive Gaussian primitive allocation.
	
	\item We exploit a high-frequency prior to devise a frequency prior guidance module and a routing regularization term, thereby ensuring a more complexity-aware and structurally sound allocation.
	
	\item Our SplatWeaver allocates Gaussian primitives in a more principled manner, leading to high-fidelity reconstructions that significantly outperform alternative methods across a variety of benchmarks.
\end{itemize}

The remaining part of this paper is organized as follows: Section \ref{2} reviews existing novel View synthesis methods and summarizes the relevant dynamic neural networks. Section \ref{3} presents the methodology of how to achieve adaptive Gaussian allocation through a dedicated cardinality Gaussian expert routing paradigm. Section \ref{4} demonstrates experiments to verify the performance of SplatWeaver on various scenarios. Lastly, Section \ref{5} provides concluding remarks.

\section{Related Work}\label{2}
	\subsection{Radiance Fields for Novel View Synthesis.}
	The advent of radiance field representations \cite{mildenhall2021nerf,kerbl20233d} has marked a paradigm revolution in novel view synthesis. A pivotal milestone in this domain is Neural Radiance Fields (NeRF) \cite{mildenhall2021nerf}, which introduced an implicit volumetric representation parameterized by coordinate-based neural networks. The success of NeRF and its variants \cite{barron2021mip,barron2022mip,barron2023zip,verbin2022ref,chen2022tensorf,fridovich2022plenoxels,garbin2021fastnerf,muller2022instant} have catalyzed a surge of research extending radiance fields to dynamic scenes \cite{park2021nerfies,park2021hypernerf,wang2023masked,fang2022fast,liu2023robust,guo2023forward,shao2023tensor4d}.
	Despite these advances, NeRF-based methods remain hampered by expensive training and slow rendering, limiting their broader practical applications. More recently, 3D Gaussian Splatting (3DGS) \cite{kerbl20233d} introduced an explicit and efficient Gaussian-based scene representation, dramatically accelerating rendering while maintaining high visual fidelity. Building upon this representation, numerous subsequent works \cite{yu2024mip,cheng2024gaussianpro,wu20244d,lu2024scaffold,tang2023dreamgaussian,zhou2025gps,fang2025efficient} have extended 3DGS to a wide range of scenarios. For instance, Mip-Splatting improves the anti-aliasing capability of 3DGS, while Scaffold-GS \cite{lu2024scaffold} achieves enhanced rendering quality through anchor-based learning. GIR \cite{shi2025gir} investigates inverse rendering for scene factorization, and StylizedGS \cite{zhang2025stylizedgs} enables controllable scene stylization.
Nevertheless, these methods typically require scene-specific optimization, which can take from several minutes to hours. In addition, they often rely on auxiliary tools, such as SfM, to estimate camera poses and initialize the scene point cloud, further limiting their applicability in real-world, in-the-wild scenarios.
    
\subsection{Generalizable Novel View Synthesis.}
Generalizable novel view synthesis \cite{chen2021mvsnerf,wang2022attention,charatan2024pixelsplat,chen2024mvsplat,ziwen2025long,jiang2025anysplat,xu2023wavenerf} has emerged as a central topic in 3D reconstruction, aiming to eliminate costly scene-specific optimization.
Early methodologies \cite{charatan2024pixelsplat,chen2024mvsplat,xu2025depthsplat,zhang2024gs,min2024epipolar,tang2024hisplat} primarily focused on reconstructing small-scale scenes from sparse observations with known camera poses. However, scenarios involving only 2–4 posed views are uncommon in real-world applications, and these methods often suffer from substantial memory overhead when handling a larger number of viewpoints due to the reliance on cost volumes.
Subsequent efforts \cite{ziwen2025long,wang2025zpressor} have extended the range of input views, enabling generalization across wider baseline configurations. Nevertheless, their dependence on a priori camera parameters restricts their utility in in-the-wild settings, particularly in unconstrained scenarios where calibration data is noisy or unavailable.
More recently, several pioneering works \cite{jiang2025anysplat,ye2026yonosplat,ye2024no,zhang2025flare,hong2024pf3plat,smart2024splatt3r} have explored the joint estimation of camera poses and scene appearance, demonstrating promising generalization capabilities and high-fidelity rendering quality.
Despite these advances, existing approaches predominantly rely on either pixel-aligned \cite{charatan2024pixelsplat,ziwen2025long,chen2024mvsplat,xu2025depthsplat} or voxel-aligned \cite{miao2025evolsplat,wang2025volsplat,jiang2025anysplat,li2026tokensplat,liu2025worldmirror} Gaussian prediction schemes, which often lead to redundancy in smooth regions and deficiency in complex areas. Although a line of work \cite{ziwen2025long,moreau2025off,singhalgaussiantrim3r,ye2026yonosplat,zhang2024gaussian,park2025ecosplat} focuses on pruning strategies to mitigate redundancy, they still fail to adaptively allocate Gaussian primitives according to scene complexity.
While recent token-query architectures, such as C3G \cite{an2025c3g} and TokenGS \cite{tokengs2026}, attempt to decouple Gaussian prediction from rigid grids, their reliance on a predefined Gaussian budget inherently constrains their adaptive scalability across diverse scenes and varying levels of view coverage.
In contrast to existing methods, we introduce the cardinality Gaussian routing paradigm that adaptively allocates Gaussian primitives based on scene complexity under a flexible budget, yielding superior rendering quality and improved efficiency for generalizable novel view synthesis.

\subsection{Dynamic Neural Networks.}
Dynamic neural networks \cite{wang2018skipnet,veit2018convolutional,jia2016dynamic,dai2017deformable,zhou2019spatio,gao2019deformable,su2016leaving,wu2019adaframe} are intended to adaptively adjust their weights or structure to handle given input with appropriate states, offering a more flexible alternative to static architectures. Recently, this paradigm has evolved from basic conditional computation \cite{bengio2013estimating} toward sophisticated Mixture-of-Experts (MoE) architectures, which effectively scale model capacity while preserving efficiency \cite{puigcerver2023sparse,riquelme2021scaling,shazeer2017outrageously}.
By employing dynamic routing, the network can better capture the diversity and heterogeneity of the data distribution. This scheme has proven successful across various vision tasks, including large multimodal models \cite{li2025uni,shen2024mome}, medical image segmentation \cite{wei2025mixture,wang2024sam}, and image restoration \cite{zamfir2025complexity,lin2024unirestorer}, etc.
In this work, we introduce the concept of cardinality Gaussian experts, where a specialized suite of experts is designed to predict varying quantities of Gaussian primitives. Through pixel-level dynamic routing, our framework enables flexible and adaptive Gaussian allocation in a feed-forward manner.

\section{Methodology}\label{3}
\begin{figure*}[t]
	\begin{center}
		\includegraphics[width=\linewidth]{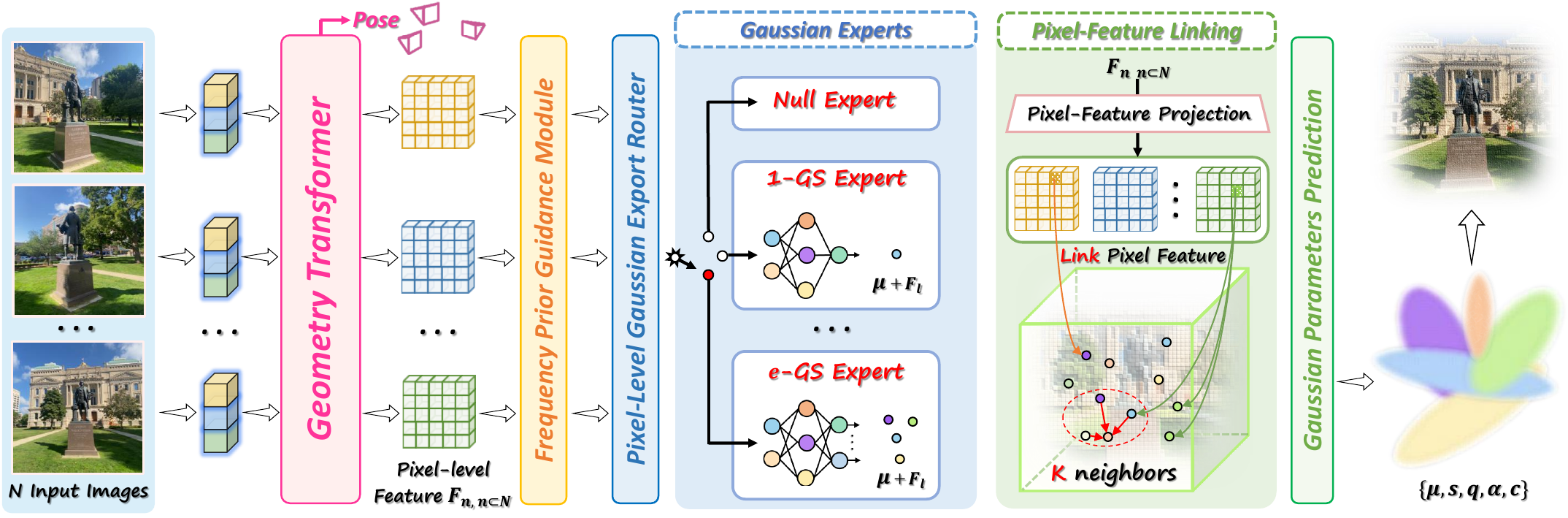}
	\end{center}
	\vspace{-7pt}
	\caption{ \textbf{Overall framework of SplatWeaver}. Given $N$ uncalibrated images, a geometry transformer first estimates camera poses and extracts pixel-level features $\{F_n\}_{n=1}^N$. Subsequently, guided by a frequency prior injection module, a router assigns each pixel to the most suitable cardinality Gaussian expert $E_e$, which predicts a set of hidden Gaussians comprising spatial positions $\mu$ and latent features $F_l$. After gathering all hidden Gaussians, we integrate their corresponding pixel-level features and aggregate neighborhood context to predict the remaining parameters for each primitive.}
	\label{figure2}
	\vspace{-10pt}
\end{figure*}

Our core insight is to adaptively allocate Gaussian primitives according to scene complexity, instead of predicting a uniform number of per-pixel or per-voxel Gaussians, thereby avoiding redundancy in simple regions and deficiency in complex areas. In particular, we advocate the concept of cardinality Gaussian experts, where each expert is responsible for predicting a specific number of Gaussian primitives (ranging from $0$ to $M$). Allocation across regions is then achieved via pixel-level cardinality Gaussian expert routing. This paradigm provides the desired flexibility, enabling the model to adapt the distribution of Gaussian primitives to the complexity of different spatial regions, while also dynamically controlling the overall budget according to the complexity and span of the entire scene. As a result, it achieves a more efficient and expressive 3D representation. The schematic illustration of the proposed SplatWeaver is depicted in Fig. \ref{figure2}.

\subsection{Preliminaries}
\noindent\textbf{Problem Formulation.} Consider $N$ uncalibrated views of a single 3D scene, given as images $\{I_n\}_{n=1}^N$, where $I_n \in \mathbb{R}^{H \times W \times 3}$, generalizable 3D Gaussian splatting models aim to
jointly recover the scene's geometry, appearance, and camera poses. Specifically, the 3D scene is represented by a collection of $G$ anisotropic 3D Gaussians:
\begin{equation}
	\mathcal{G} = \{(\boldsymbol{\mu}^{(g)}, \boldsymbol{s}^{(g)}, \boldsymbol{q}^{(g)}, \alpha^{(g)}, \boldsymbol{c}^{(g)})\}_{g=1}^G,
\end{equation}
where each Gaussian is parameterized by its mean position $\boldsymbol{\mu} \in \mathbb{R}^3$, an anisotropic scaling factor $\boldsymbol{s} \in \mathbb{R}^3$, a rotation quaternion $\boldsymbol{q} \in \mathbb{R}^4$, an opacity value $\alpha \in \mathbb{R}^+$, and a color embedding $\boldsymbol{c} \in \mathbb{R}^{3 \times (k+1)^2}$ represented via spherical harmonic (SH) coefficients of degree $k$. Simultaneously, the model estimates the camera parameters for each view:
\begin{equation}
	\mathcal{P} = \{p_n \in \mathbb{R}^9\}_{n=1}^N,
\end{equation}
where $p_n$ encapsulates both the intrinsic and extrinsic parameters of the $n$-th view. Formally, our model learns a mapping $f_{\theta}$ that predicts the 3D primitives and camera poses directly from the input images:
\begin{equation}
	f_{\theta}: \{I_n\}_{n=1}^N \longmapsto \mathcal{G} \cup \mathcal{P}.
\end{equation}

\subsection{Overview of SplatWeaver}
Given $N$ uncalibrated images $\{I_n\}_{n=1}^N$, where $I_n \in \mathbb{R}^{H \times W \times 3}$, SplatWeaver first patchifies each image $I_n$ into $\frac{H\times W}{p^2}$ tokens using DINOv2 \cite{oquab2023dinov2}. It then incorporates a multi-view geometry transformer to extract interactive features and predict camera pose parameters $p_n$, following the principles of VGGT \cite{Wang_2025_CVPR}. Subsequently, a DPT-like decoder \cite{ranftl2021vision} is utilized to obtain the pixel-level per-image features $\{F_n\}_{n=1}^N$, where $F_n \in \mathbb{R}^{H \times W \times D}$.
To ensure robust routing, a frequency prior guidance module is employed to extract high-frequency priors from the discrete wavelet domain, which guides the network toward more reliable expert allocations. Following this, a pixel-level Gaussian expert router assigns the most appropriate cardinality Gaussian expert $E_e$ to each pixel-wise feature. Each expert is tasked with predicting a specific number of hidden Gaussians, yielding their spatial positions $\mu$ and latent features $F_l$. These predicted hidden Gaussians are then concatenated with their corresponding projected pixel features $F_n^p(i,j)$ (where $i \in H, j \in W$) to construct the combined representation $\{\mu^{(g)}, F_l^{(g)}, F_n^p(i,j)\}$. Finally, by leveraging the features of $K$ neighboring hidden Gaussians, the framework predicts the remaining attributes of each Gaussian primitive via attention-based aggregation, including its scale $s^{(g)}$, rotation $q^{(g)}$, opacity $\alpha^{(g)}$, and color $c^{(g)}$.

\subsection{Cardinality Gaussian Expert Routing}

To enable adaptive Gaussian allocation in feed-forward 3D reconstruction, we introduce the concept of cardinality Gaussian experts, where each expert is responsible for predicting a specific number of Gaussian primitives. By dynamically routing specific experts to different spatial regions according to scene content and geometry, this approach guarantees a flexible and complexity-aware distribution of Gaussian primitives.

\noindent\textbf{Cardinality Gaussian Expert.} Instead of requiring experts to predict all Gaussian parameters directly, which would result in a lack of spatial context awareness and suboptimal prediction quality, we advocate that each expert predicts only the Gaussian positions and their corresponding latent features. The remaining parameters are then decoded with enhanced precision by leveraging the surrounding spatial context, as elaborated in the next section.
Specifically, we first deliberately introduce the null expert that predicts no Gaussian primitives, thereby enabling sparsity and flexibility in Gaussian allocation.
Each remaining expert $E_e$ is implemented as a lightweight predictor composed of two linear layers with a ReLU activation function.
Given a pixel-wise feature $F_n(i,j)$, the expert predicts a set of hidden Gaussian primitives characterized by their positions and latent features:
\begin{equation}
	\setlength{\abovedisplayskip}{3pt}
	\setlength{\belowdisplayskip}{3pt}
	\begin{aligned}
		\{\mu^{(g)}, F_l^{(g)}\}_{g=1}^{m_e} = E_e\!\left(F_n(i,j)\right),
	\end{aligned}
\end{equation}
where $\mu^{(g)} \in \mathbb{R}^3$ denotes the 3D position of the $g$-th hidden Gaussian primitive, $F_l^{(g)} \in \mathbb{R}^{d}$ represents its latent feature. $m_e \in \{0, 1, \dots, M\}$ indicates the cardinality associated with expert $E_e$, i.e., the number of Gaussian primitives predicted by that expert.
$M$ is empirically set to 3, i.e., an expert predicts at most three Gaussian primitives. We found that this cardinality not only ensures fine-grained scene representation but also balances Gaussian prediction reliability and routing complexity. 

\begin{figure*}[t]
	\begin{center}
		\includegraphics[width=\linewidth]{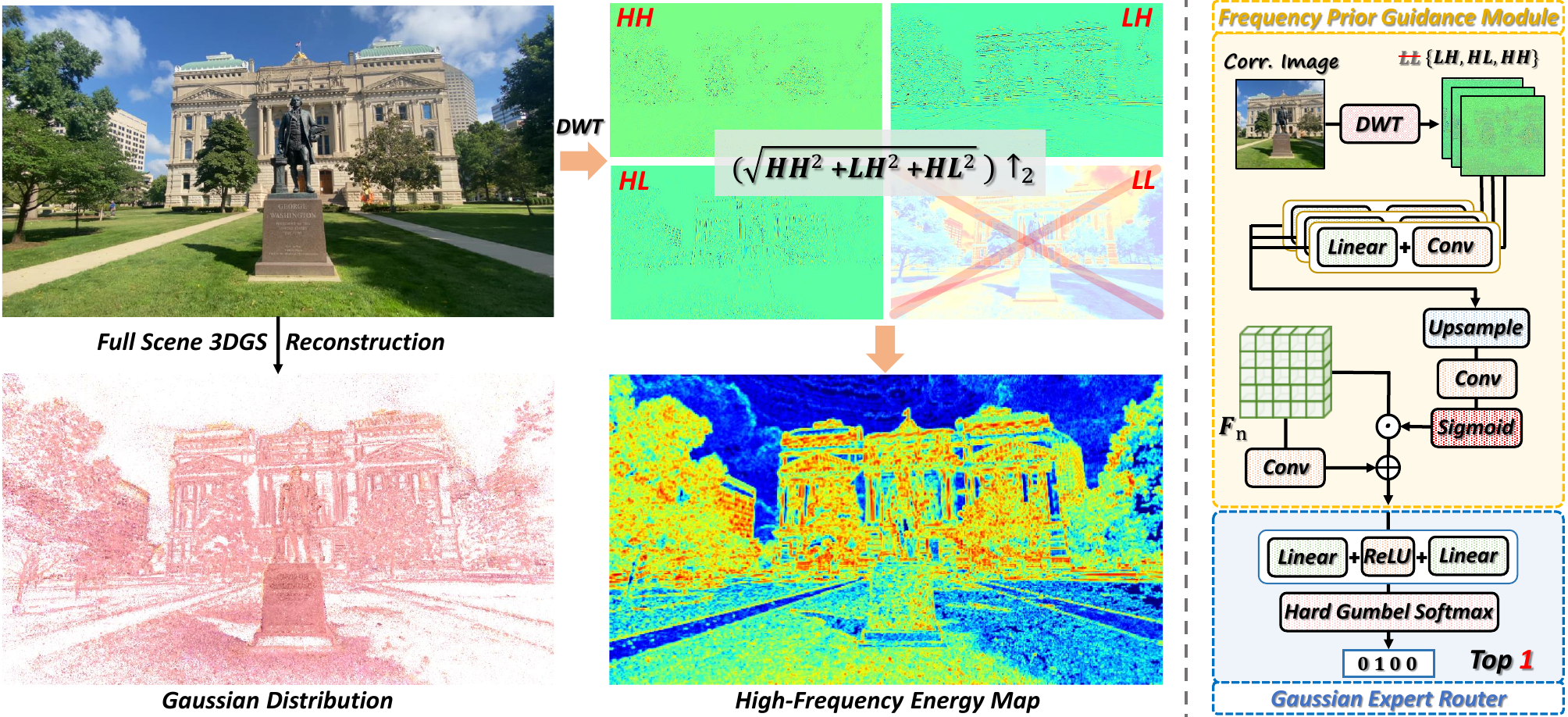}
	\end{center}
	\vspace{-7pt}
	\caption{ \textbf{Left:} Illustration of the proposed high-frequency prior, where the high-frequency energy map, derived from the discrete wavelet transform with $\scriptstyle(\sqrt{\mathrm{HH}^2+\mathrm{LH}^2+\mathrm{HL}^2}) \uparrow_{2}$, exhibits strong alignment with the Gaussian distribution obtained from full scene reconstruction via 3DGS. \textbf{Right:} Diagram of the proposed frequency prior guidance module and the pixel-level Gaussian expert router.}
	\label{figure3}
	\vspace{-16pt}
\end{figure*} 
\noindent\textbf{Frequency Prior Guided Routing.} %
Without routing supervision or constraints, the model may struggle to learn appropriate allocations of Gaussian experts. Moreover, since the null expert does not produce gradients, their routing assignments cannot be directly optimized via the reconstruction loss item. To address this issue, as illustrated in Fig.~\ref{figure3}, we observe that the high-frequency energy map (HF) derived from the discrete wavelet transform (DWT), exhibits strong alignment with the Gaussian distribution obtained from dense scene reconstruction using 3D Gaussian Splatting (3DGS).
\begin{equation}
	\setlength{\abovedisplayskip}{2pt}
	\setlength{\belowdisplayskip}{2pt}
	\begin{aligned}
	 (L\!L, L\!H, H\!L,H\!H)=\mathrm{DWT}(I), \\
		H\!F=(\sqrt{L\!H^2+H\!L^2+H\!H^2}) ~ \scalebox{1.2}{$\uparrow$}_{2}, 
	\end{aligned}
\end{equation} 
where $\uparrow_2$ denotes an upsampling operator with a scale factor of 2.
This dense reconstruction serves as a valuable reference for Gaussian allocation. It is intuitive that regions with high-frequency energy typically correspond to areas rich in structural detail, which necessitate a higher density of Gaussian primitives to model fine-grained scene content. Consequently, this characteristic can serve as an ideal auxiliary prior for guiding expert selection.
In practice, we introduce a frequency prior guidance module to inject frequency prior into the feature representation and design a dedicated routing regularization term based on the high-frequency energy map to guide expert allocation. The details are elaborated below.

\noindent\textbf{Frequency Prior Guidance Module.} The frequency prior guidance module serves as a precursor to the expert router, specifically designed to enrich pixel-level features with complexity-aware information. 
As illustrated in Fig. \ref{figure3}, for the pixel-level features $F_n$ of a given view, we first apply a Discrete Wavelet Transform (DWT) to the corresponding input image to extract high-frequency components, denoted as $\scriptstyle \{L\!H, H\!L, H\!H\}$. 
These components are processed through parallel branches consisting of linear and convolutional layers. 
Subsequently, the features are passed through an upsampling layer and a final convolutional block to restore the spatial dimensions. 
Finally, a sigmoid activation function is employed to generate a frequency-aware attention map, which is then used to modulate the original features $F_n$. This process can be formulated as:
\begin{equation}
	F_n^f = F_n \odot \sigma(\Psi(\{L\!H, H\!L, H\!H\})) + \text{Conv}(F_n),
\end{equation}
where $\Psi$ denotes the series of transformation layers, $\sigma$ is the Sigmoid function, and $\odot$ represents element-wise multiplication. 
This mechanism effectively guides the expert router to prioritize regions with high structural complexity by modulating the feature representation.

\noindent\textbf{Pixel-Wise Expert Router.} As depicted in Fig. \ref{figure3}, the expert router $R$ is implemented using two linear layers with a ReLU activation function. 
Given the pixel-wise feature from the frequency prior guidance module $F_n^f(i,j)$, the router predicts routing logits over all experts. 
To obtain discrete routing decisions while maintaining differentiability, we employ the Gumbel-Softmax trick with the Straight-Through estimator to generate a one-hot routing probability:
\begin{equation}
	\begin{aligned}
		p_n(i,j) = \mathrm{GumbelSoftmax}(R(F_n^f(i,j))),
	\end{aligned}
\end{equation}
where $p_n(i,j)$ is a one-hot vector indicating the selected expert for pixel $(i,j)$. During routing, we select the top-1 cardinality Gaussian expert according to $p_n(i,j)$ for prediction.
The output of the selected expert is multiplied by the corresponding routing probability. 
Since $p_n(i,j)$ is a hard one-hot vector, the weighting factor is $1$ for the chosen expert, thereby preserving the physical meaning of the expert's spatial predictions (e.g., $\mu$), which would otherwise be compromised by soft probability weighting.

\begin{figure}[t]
	\begin{center}
		\includegraphics[width=.85\linewidth]{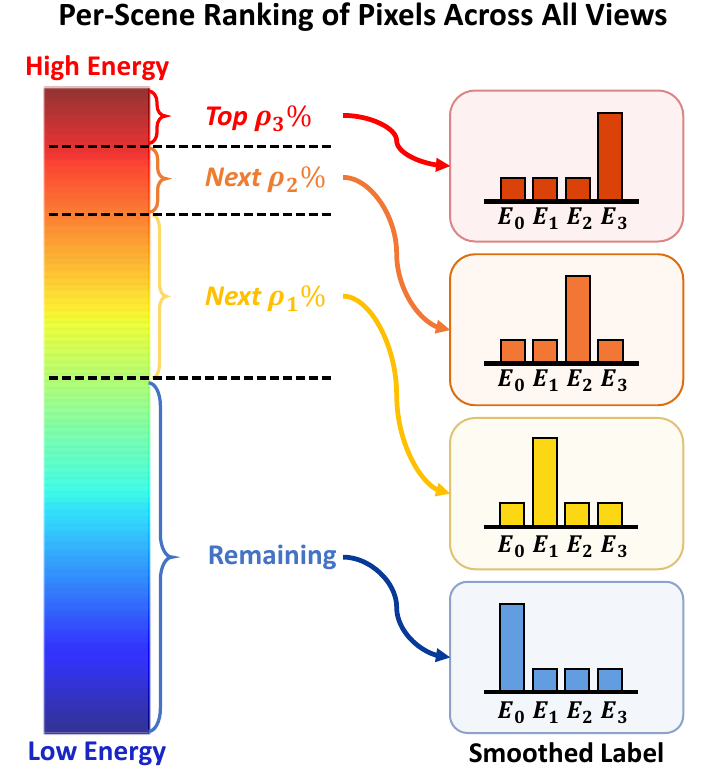}
	\end{center}
	\vspace{-7pt}
	\caption{Diagram of the proposed frequency prior-guided routing regularization scheme.}
	\label{figure_reg}
	\vspace{-11pt}
\end{figure} 

\noindent\textbf{Routing Regularization.}
To further stabilize expert routing using the aforementioned frequency prior, we introduce an auxiliary routing regularization loss derived from the high-frequency energy map computed from the discrete wavelet coefficients (Fig. \ref{figure_reg}).
Intuitively, pixels with higher frequency energy typically correspond to more complex structures, and vice versa. Therefore, we rank all pixels according to their energy values and assign routing supervision accordingly. 
Pixels with higher energy are encouraged to route to experts with larger cardinalities, while pixels with lower energy are encouraged to select experts with smaller cardinalities. 
Concretely, the top $\rho_3\%$ pixels are assigned to the expert $E_3$, the next $\rho_2\%$ pixels are assigned to the expert $E_2$, and the following $\rho_1\%$ pixels are assigned to the expert $E_1$, and the remaining pixels are assigned to the expert $E_0$. 
This assignment serves as a soft supervision signal for routing, implemented via a cross-entropy loss with label smoothing ($\epsilon=0.1$).:
\begin{equation}
	\setlength{\abovedisplayskip}{1pt}
	\setlength{\belowdisplayskip}{-1pt}
	\begin{aligned}
 \tilde{y}_n^{(e)}(i,j) = (1 - \epsilon) y_n^{(e)}(i,j) + \frac{\epsilon}{E}, \\
 	\mathcal{L}_{\text{route}}
 = -\sum\nolimits_{i,j} \sum\nolimits_{e} \tilde{y}_n^{(e)}(i,j)
 \log p_n^{(e)}(i,j), 
 	\end{aligned}
\end{equation}
where $p_n^{(e)}(i,j)$ denotes the routing probability for expert $E_e$ at pixel $(i,j)$, and $\tilde{y}_n^{(e)}(i,j)$ represents the smoothed routing target derived from the energy-based ranking labels $y_{n}^{(e)}(i,j) \in \{0, 1\}$. Wherein $E$ is the total number of experts and $\epsilon$ is the smoothing factor. 

Notably, this routing regularization is only applied during the first half of training to guide reasonable expert allocation. 
In the later training stage, the constraint is removed, leaving only the budget term to allow the model to autonomously explore the optimal routing strategy:
\begin{equation}
	\begin{aligned}
		\mathcal{L}_{\text{budget}} = \max(0, G - \epsilon N\!H\!W)^2,
	\end{aligned}
\end{equation}
This constraint penalizes the model only when the total number of predicted Gaussian primitives $G$ exceeds $\epsilon N\!H\!W$ ($\epsilon=0.3$ by default), thereby encouraging a compact and efficient representation.

\begin{figure}[t]
	\begin{center}
		\includegraphics[width=\linewidth]{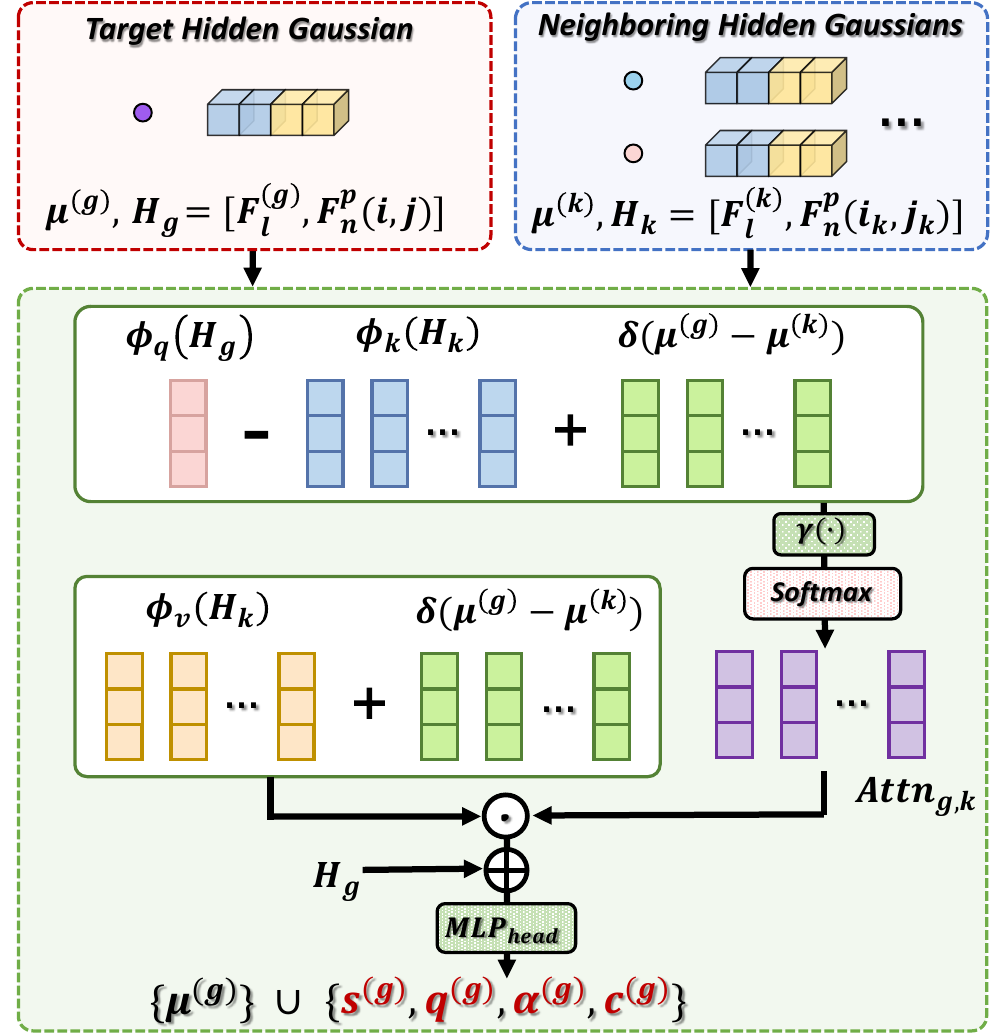}
	\end{center}
	\vspace{-7pt}
	\caption{Illustration of the proposed feature aggregation and parameter prediction module.}
	\label{figure_neighbor}
	\vspace{-11pt}
\end{figure} 
\subsection{Neighbor-Conditioned Gaussian Parameter Prediction}
Instead of predicting the final Gaussian attributes in isolation, this module leverages local neighboring hidden Gaussians to refine their parameters through an attention-based aggregation scheme.

As previously discussed, rather than predicting complete Gaussian primitive parameters directly, we cultivate experts to output a set of latent features $F_l$. To achieve more precise and context-aware attribute estimation, we subsequently aggregate the features of spatial neighbors from the predicted ``hidden Gaussians.'' 
This design allows the model to leverage local geometric consistency and spatial context, ensuring that the final Gaussian parameters are physically coherent and better aligned with the underlying scene structure.

\noindent\textbf{Pixel-Feature Linking.} To preserve the initial pixel-level detail-rich features, we associate each hidden Gaussian $\{\mu^{(g)}, F_l^{(g)}\}$ with its corresponding original pixel feature, forming the triplet $\{\mu^{(g)}, F_l^{(g)}, F_n^p(i,j)\}$, where $F_n^p(i,j)$ denotes the projected feature of $F_n(i,j)$ obtained via a MLP projection layer:
\begin{equation}
	F_n^p(i,j) =  MLP(F_n(i,j)).
\end{equation}

\noindent\textbf{Feature Aggregation and Parameter Prediction.} Given the hidden Gaussian representations, we establish spatial context by aggregating features from local neighbors. Due to the massive number of Gaussian primitives, performing direct k-nearest neighbor (KNN) matching entails significant computational overhead. To this end, we leverage the Faiss CUDA acceleration library \cite{johnson2019faiss} and adopt a coarse-to-fine strategy, i.e., performing clustering followed by local matching, which achieves millisecond-level $8$ nearest neighbor search across millions of candidates. 

\begin{table*}[t]
	\setlength{\tabcolsep}{1.3pt}	
	\caption{Quantitative comparisons novel view synthesis on DL3DV \cite{ling2024dl3dv}, RealEstate10K \cite{zhou2018stereo}, and Mip-NeRF 360 \cite{barron2022mip}. We evaluate all models with
		4, 8, 16, and 24 input views. $^\dagger$ denotes the extreme-efficiency variant with the fewest Gaussians and competitive performance. }
	\footnotesize
	\vspace{-7pt}
	\begin{center}
		\begin{tabular}{l|cccc|cccc|cccc|cccc}
			\toprule[1pt]
			\makecell[l]{\multirow{2}{*}{\textbf{Method}}} &\multicolumn{4}{c|}{\textbf{4 Views}}&\multicolumn{4}{c|}{\textbf{8 Views}}&\multicolumn{4}{c|}{\textbf{16 Views}}&\multicolumn{4}{c}{\textbf{24 Views}}
			\\
			& PSNR$\uparrow$ & SSIM$\uparrow$&  LPIPS$\downarrow$&GS$_\times10^3$& PSNR$\uparrow$ & SSIM$\uparrow$&  LPIPS$\downarrow$&GS$_\times10^3$& PSNR$\uparrow$ & SSIM$\uparrow$&  LPIPS$\downarrow$&GS$_\times10^3$& PSNR$\uparrow$ & SSIM$\uparrow$&  LPIPS$\downarrow$&GS$_\times10^3$\\
			\midrule[1pt]
			\multicolumn{17}{c}{DL3DV}\\
			\midrule
			NoPoSplat \cite{ye2024no}&14.02&0.367&0.651&458&13.92&0.370&0.653&918&13.88&0.367&0.651&1835&13.81&0.365&0.659&2710\\
			FLARE \cite{zhang2025flare}&13.52&0.348&0.674&458&13.55&0.352&0.682&918&13.44&0.339&0.686&1835&13.37&0.332&0.695&2710\\
			SPFSplat \cite{huang2025no} &15.47&0.439&0.599&458&16.48&0.443&0.512&918&17.22&0.473&0.438&1835&18.02&0.502&0.401&2710\\
			YoNoSplat \cite{ye2026yonosplat}&16.19&0.441&0.442&312&17.39&0.456&0.427&569&17.63&0.466&0.429&1259&18.48&0.488&0.424&1728\\
			AnySplat \cite{jiang2025anysplat} &15.61&0.423&0.388&413&17.75&0.473&0.336&805&19.09&0.558&0.281&1522&19.93&0.601&0.266&2173\\
			AnySplat \cite{jiang2025anysplat}+\cite{fan2024lightgaussian} &6.24&0.105&0.711&184&7.42&0.125&0.699&367&9.25&0.169&0.672&734&12.37&0.224&0.631&1084\\
			EcoSplat$_{40\%}$ \cite{park2025ecosplat} &13.41&0.310&0.629&184&13.96&0.386&0.646&367&14.64&0.403&0.635&734&15.57&0.433&0.602&1084\\
			C3G \cite{an2025c3g}&9.41&0.167&0.715&2&9.90&0.202&0.701&2&10.48&0.224&0.710&2&10.72&0.244&0.706&2\\
			SplatWeaver$^\dagger$ &15.97&0.438&0.397&44&17.99&0.518&0.324&89&19.52&0.587&0.279&153&20.57&0.614&0.258&211\\
			\textbf{SplatWeaver} &\cellcolor{blue!10}\textbf{16.67}&\cellcolor{blue!10}\textbf{0.473}&\cellcolor{blue!10}\textbf{0.375}&\cellcolor{blue!10}128&\cellcolor{blue!10}\textbf{18.68}&\cellcolor{blue!10}\textbf{0.546}&\cellcolor{blue!10}\textbf{0.312}&\cellcolor{blue!10}238&\cellcolor{blue!10}\textbf{20.11}&\cellcolor{blue!10}\textbf{0.607}&\cellcolor{blue!10}\textbf{0.260}&\cellcolor{blue!10}451&\cellcolor{blue!10}\textbf{21.04}&\cellcolor{blue!10}\textbf{0.626}&\cellcolor{blue!10}\textbf{0.251}&\cellcolor{blue!10}548\\
			\midrule[1pt]
			\multicolumn{17}{c}{RealEstate10K}\\
			\midrule
			NoPoSplat \cite{ye2024no}&16.75&0.539&0.433&458&16.36&0.525&0.457&918&16.06&0.467&0.513&1835&15.82&0.411&0.560&2710\\
			FLARE \cite{zhang2025flare}&14.29&0.471&0.557&458&14.92&0.470&0.563&918&14.11&0.455&0.603&1835&13.97&0.432&0.617&2710\\
			SPFSplat \cite{huang2025no} &18.05&0.601&0.358&458&18.46&0.687&0.285&918&18.94&0.701&0.266&1835&19.58&0.719&0.258&2710\\
			YoNoSplat \cite{ye2026yonosplat}&19.21&0.612&0.328&318&19.82&0.635&0.324&580&19.73&0.656&0.326&1238&20.11&0.649&0.322&1430\\
			AnySplat \cite{jiang2025anysplat} &19.02&0.622&0.297&384&20.86&0.694&0.237&728&22.28&0.744&0.201&1312&23.15&0.788&0.178&1781\\
			AnySplat \cite{jiang2025anysplat}+\cite{fan2024lightgaussian} &8.63&0.291&0.611&184&11.21&0.452&0.572&367&13.26&0.501&0.539&734&15.66&0.550&0.499&1084\\
			EcoSplat$_{40\%}$ \cite{park2025ecosplat} &15.73&0.515&0.488&184&17.19&0.555&0.470&367&18.45&0.572&0.354&734&19.38&0.617&0.333&1084\\
            C3G \cite{an2025c3g}&12.03&0.415&0.585&2&12.07&0.418&0.581&2&12.13&0.421&0.580&2&12.32&0.430&0.579&2\\
			SplatWeaver$^\dagger$ &19.03&0.624&0.289&39&21.22&0.717&0.214&82&22.75&0.762&0.189&142&23.66&0.799&0.170&197\\
			\textbf{SplatWeaver} &\cellcolor{blue!10}\textbf{19.35}&\cellcolor{blue!10}\textbf{0.635}&\cellcolor{blue!10}\textbf{0.274}&\cellcolor{blue!10}113&\cellcolor{blue!10}\textbf{21.47}&\cellcolor{blue!10}\textbf{0.728}&\cellcolor{blue!10}\textbf{0.204}&\cellcolor{blue!10}218&\cellcolor{blue!10}\textbf{22.96}&\cellcolor{blue!10}\textbf{0.784}&\cellcolor{blue!10}\textbf{0.182}&\cellcolor{blue!10}417&\cellcolor{blue!10}\textbf{23.85}&\cellcolor{blue!10}\textbf{0.815}&\cellcolor{blue!10}\textbf{0.164}&\cellcolor{blue!10}524\\
			\midrule[1pt]
			
			\multicolumn{17}{c}{Mip-NeRF 360}\\
			\midrule
			NoPoSplat \cite{ye2024no}&14.03&0.259&0.682&458&13.74&0.258&0.715&918&13.59&0.255&0.732&1835&13.28&0.257&0.729&2710\\
			FLARE \cite{zhang2025flare}&14.52&0.279&0.670&458&13.97&0.255&0.690&918&13.72&0.250&0.722&1835&13.48&0.242&0.731&2710\\
			SPFSplat \cite{huang2025no} &14.65&0.266&0.633&458&14.24&0.270&0.668&918&13.79&0.265&0.695&1835&14.27&0.275&0.670&2710\\
			YoNoSplat \cite{ye2026yonosplat}&14.29&0.268&0.634&377&14.53&0.272&0.622&684&14.54&0.273&0.625&1350&14.61&0.272&0.618&1725\\
			AnySplat \cite{jiang2025anysplat} &12.33&0.300&0.484&435&17.15&0.430&0.349&837&18.94&0.519&0.300&1628&19.55&0.534&0.290&2141\\
			AnySplat \cite{jiang2025anysplat}+\cite{fan2024lightgaussian} &5.38&0.057&0.739&184&6.67&0.087&0.725&367&8.87&0.149&0.722&734&11.77&0.208&0.692&1084\\
			EcoSplat$_{40\%}$ \cite{park2025ecosplat} &13.19&0.235&0.719&184&12.96&0.209&0.728&367&13.16&0.245&0.728&734&13.42&0.251&0.699&1084\\
			C3G \cite{an2025c3g}&8.97&0.200&0.755&2&9.05&0.174&0.762&2&9.12&0.167&0.758&2&9.04&0.171&0.757&2\\
			SplatWeaver$^\dagger$ &14.22&0.297&0.516&48&17.09&0.435&0.358&97&19.31&0.522&0.301&164&20.15&0.543&0.278&224\\
			\textbf{SplatWeaver} &\cellcolor{blue!10}\textbf{15.38}&\cellcolor{blue!10}\textbf{0.355}&\cellcolor{blue!10}\textbf{0.452}&\cellcolor{blue!10}135&\cellcolor{blue!10}\textbf{18.02}&\cellcolor{blue!10}\textbf{0.473}&\cellcolor{blue!10}\textbf{0.303}&\cellcolor{blue!10}250&\cellcolor{blue!10}\textbf{20.15}&\cellcolor{blue!10}\textbf{0.552}&\cellcolor{blue!10}\textbf{0.270}&\cellcolor{blue!10}469&\cellcolor{blue!10}\textbf{20.87}&\cellcolor{blue!10}\textbf{0.571}&\cellcolor{blue!10}\textbf{0.262}&\cellcolor{blue!10}589\\
			
			\bottomrule[1pt]
		\end{tabular}
	\end{center}
	\label{table1}
	\vspace{-19pt}
\end{table*}
Whereupon, we employ a point transformer–style design \cite{zhao2021point} that computes self-attention conditioned on relative spatial positions to aggregate neighboring hidden Gaussian features. As illustrated in Fig. \ref{figure_neighbor}, for a target hidden Gaussian, we define its feature representation as:
\begin{equation}
H_g := \big[F_l^{(g)}, F_n^p(i,j)\big],
\end{equation}
Given its neighbors $k \in \text{KNN}(g)$, the attention mechanism is formulated as:
\begin{equation}
	\text{Attn}_{g,k} = \text{Softmax} (\gamma ( \phi_q(H_g) - \phi_k(H_k) + \delta(\mu^{(g)} - \mu^{(k)}) )),
\end{equation}
where $\phi_q$ and $\phi_k$ are linear projections, $\delta(\cdot)$ is a relative positional encoding MLP, and $\gamma(\cdot)$ is a attention projection MLP. The aggregated feature $\hat{H_g}$ is then computed as:
\begin{equation}
	\hat{H_g} = \sum\nolimits_{k \in \text{KNN}(g)} \text{Attn}_{g,k} \odot ( \phi_v(H_k) + \delta(\mu^{(g)} - \mu^{(k)}) ).
\end{equation}

The refined feature $\hat{H}_g$ is combined with $H_g$ via residual addition and passed through a prediction head to decode the final Gaussian attributes. 
\begin{equation}
	\{s^{(g)}, q^{(g)}, \alpha^{(g)}, c^{(g)}\} = \text{MLP}_{\text{head}}(\hat{H_g}+H_g).
\end{equation}
This approach allows the model to leverage local spatial context, ensuring that the predicted primitives are physically coherent and well-aligned with the scene structure.

\subsection{Training Objective}
Drawing inspiration from prior literature~\cite{jiang2025anysplat}, we employ a pre-trained VGGT to distill camera parameters via a Huber loss ($\mathcal{L}_{\text{pose}}$) and preserve scene geometry (i.e., depth) using a mean squared error loss ($\mathcal{L}_{\text{depth}}$). For image rendering supervision, we employ a combination of mean squared error (MSE) loss and perceptual loss between the rendered images $\{\hat{I}_n\}_{n=1}^{N}$ and the input images $\{I_n\}_{n=1}^{N}$:
\begin{equation}
	\mathcal{L}_{\text{render}} =
	\frac{1}{N}\sum\nolimits_{n=1}^{N}
	(
	\text{MSE}(I_n, \hat{I}_n)
	+ \lambda\,\text{Perceptual}(I_n, \hat{I}_n)
	),
\end{equation}
where $\lambda$ is set to 0.05.
The final training objective is defined as the weighted combination of all aforementioned loss terms:
\begin{equation}
	\mathcal{L} =
	\mathcal{L}_{\text{render}}
	+ \lambda_1 \mathcal{L}_{\text{route}}
	+ \lambda_2 \mathcal{L}_{\text{budget}}
	+ \lambda_3 \mathcal{L}_{\text{pose}}
	+ \lambda_4 \mathcal{L}_{\text{depth}},
\end{equation}
where $\lambda_1$, $\lambda_2$, $\lambda_3$, and $\lambda_4$ are empirically set to 0.01, 0.01, 10, and 0.1, respectively.

\begin{figure*}[th!]
	\begin{center}
		\includegraphics[width=\linewidth]{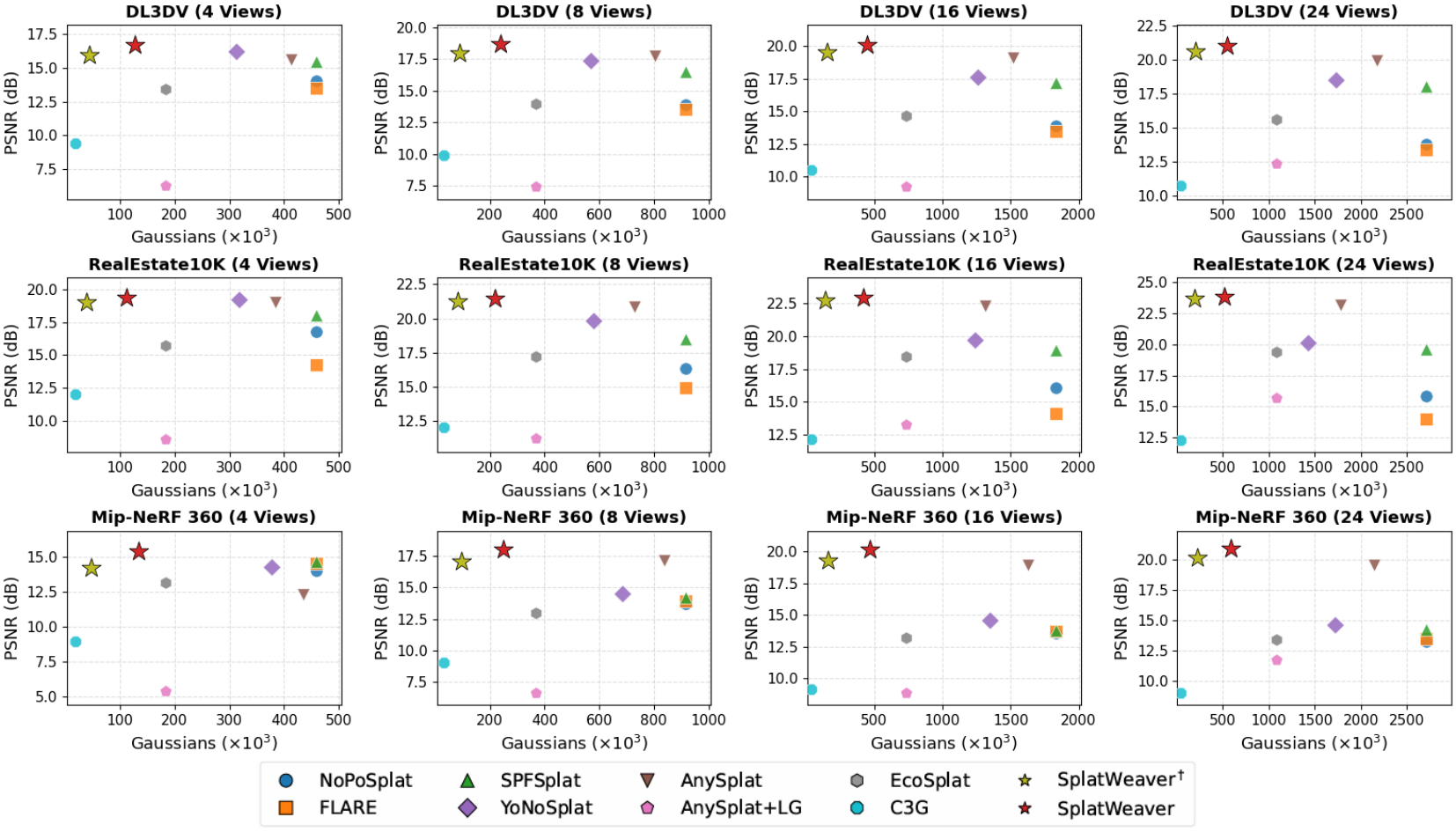}
	\end{center}
	\vspace{-7pt}
	\caption{ Comparative analysis of rendering quality versus Gaussian complexity across benchmarks under varying view settings. Our method consistently achieves superior quality–efficiency trade-offs.}
	\label{figure_scatter}
	\vspace{-10pt}
\end{figure*} 
\section{Experiments and Analysis}\label{4}
\subsection{Experimental Settings}\label{4.1}
\noindent\textbf{Implementation Details.} All experiments are implemented using the PyTorch framework and trained on eight NVIDIA A100 GPUs. We set the initial learning rate to 2e-4, which is decayed to 1e-6 following a cosine annealing schedule. During training, we randomly sample between 2 and 24 context images per batch. The maximum input resolution is limited to 448 pixels on the longer side, with the aspect ratio randomized between 0.5 and 1.0 to enhance robustness. $\rho_3\%$, $\rho_2\%$, and $\rho_1\%$ are set to $2\%$, $2\%$, and $20\%$, respectively.

We train SplatWeaver by sampling views across nine diverse public datasets: Hypersim \cite{roberts2021hypersim}, ARKitScenes \cite{baruch2021arkitscenes}, BlendedMVS \cite{yao2020blendedmvs}, ScanNet++ \cite{yeshwanth2023scannet++}, CO3D-v2 \cite{reizenstein2021common}, Objaverse \cite{deitke2023objaverse}, Unreal4K \cite{tosi2021smd}, WildRGBD \cite{xia2024rgbd}, and DL3DV \cite{ling2024dl3dv}. Our primary evaluation is conducted on the DL3DV benchmark \cite{ling2024dl3dv}, utilizing a held-out set of 140 scenes excluded from training. DL3DV benchmark encompasses a vast array of diverse environments, spanning both intricate indoor settings and expansive outdoor scenes. To further demonstrate the generalization capability of our model, we perform zero-shot evaluations on RealEstate10K \cite{zhou2018stereo} and Mip-NeRF 360 \cite{barron2022mip}. RealEstate10K test set consists of diverse indoor and outdoor scenes collected from real-world real estate videos, providing rich variations in camera motion, scene layout, and illumination conditions. Mip-NeRF 360 comprises 7 real-world scenes, including 3 outdoor and 4 indoor environments. The dataset captures unbounded $360^\circ$ scenes with complex geometries and large depth variations, posing significant challenges for view synthesis and 3D reconstruction.
For RealEstate10K, we filtered the dataset to remove 435 scenes with fewer than 32 frames due to the high overlap rate. During evaluation, we directly sample one out of every eight images throughout the entire sequence as test views, while the context views are randomly selected from the remaining frames. For DL3DV and Mip-NeRF 360, we first select 60, 80, 100, and 120 views, as well as 24, 32, 40, and 48 views, respectively, to construct the 4-view, 8-view, 16-view, and 24-view settings. Within these subsets, we designated every eighth image as a test view and utilized the remainder for context view sampling.
Due to variations in token partitioning strides among different methods, evaluations are conducted at compatible resolutions for fairness: methods with a stride of 14 are tested at $252 \times 448$, while those with a stride of 16 are evaluated at the closest matching resolution of $256 \times 448$. The implementation code will be available at \url{https://github.com/yecongwan/SplatWeaver}.

\begin{figure*}[th!]
	\begin{center}
		\includegraphics[width=.97\linewidth]{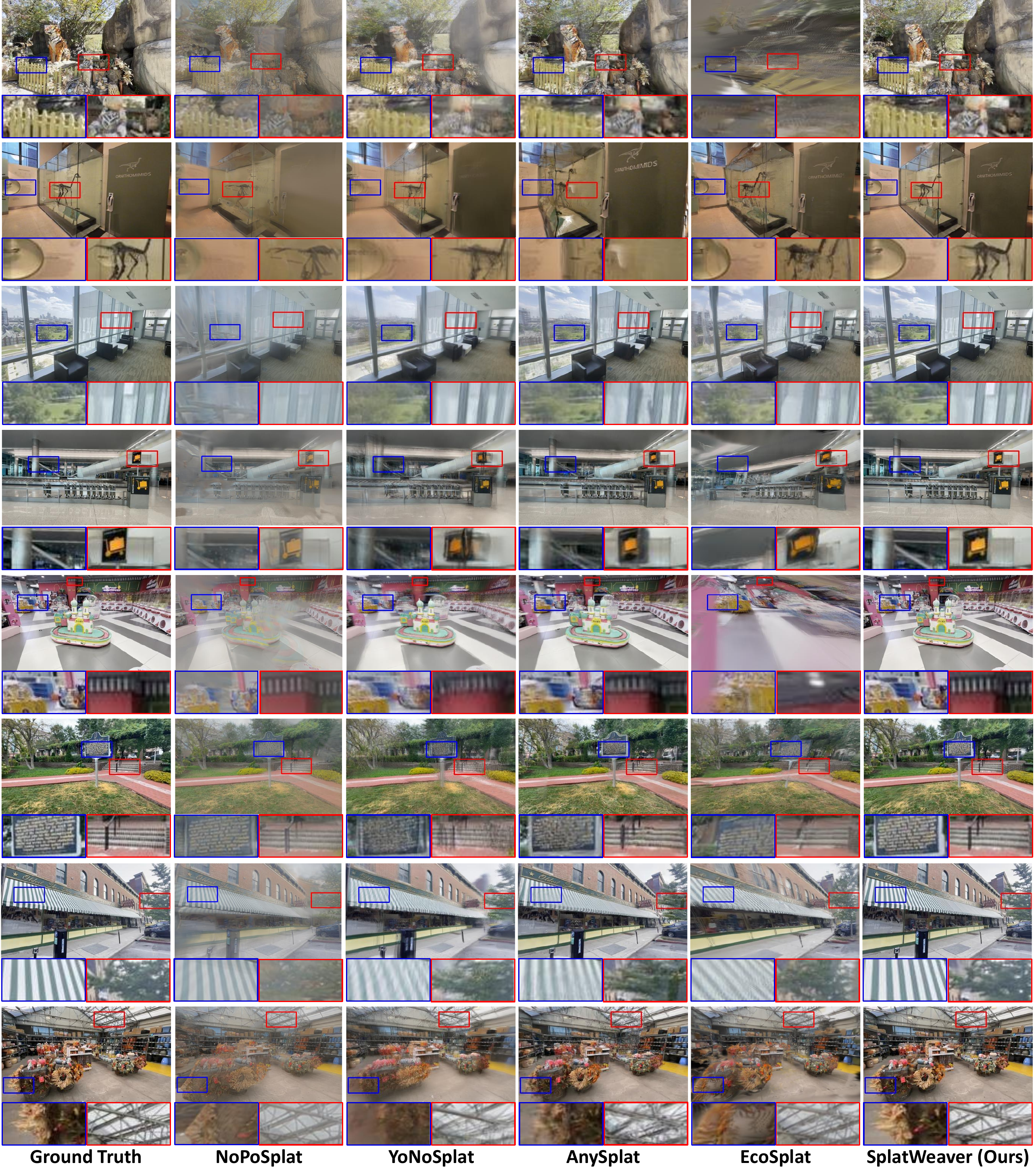}
	\end{center}
	\vspace{-7pt}
	\caption{ Qualitative comparisons on the DL3DV \cite{ling2024dl3dv} dataset. From top to bottom, every two rows correspond to rendering results under 4, 8, 16, and 24 view settings, respectively. Our method yields more coherent fine structures and sharper details.}
	\label{figure_v1}
	\vspace{-10pt}
\end{figure*} 
\begin{figure*}[th!]
	\begin{center}
		\includegraphics[width=.97\linewidth]{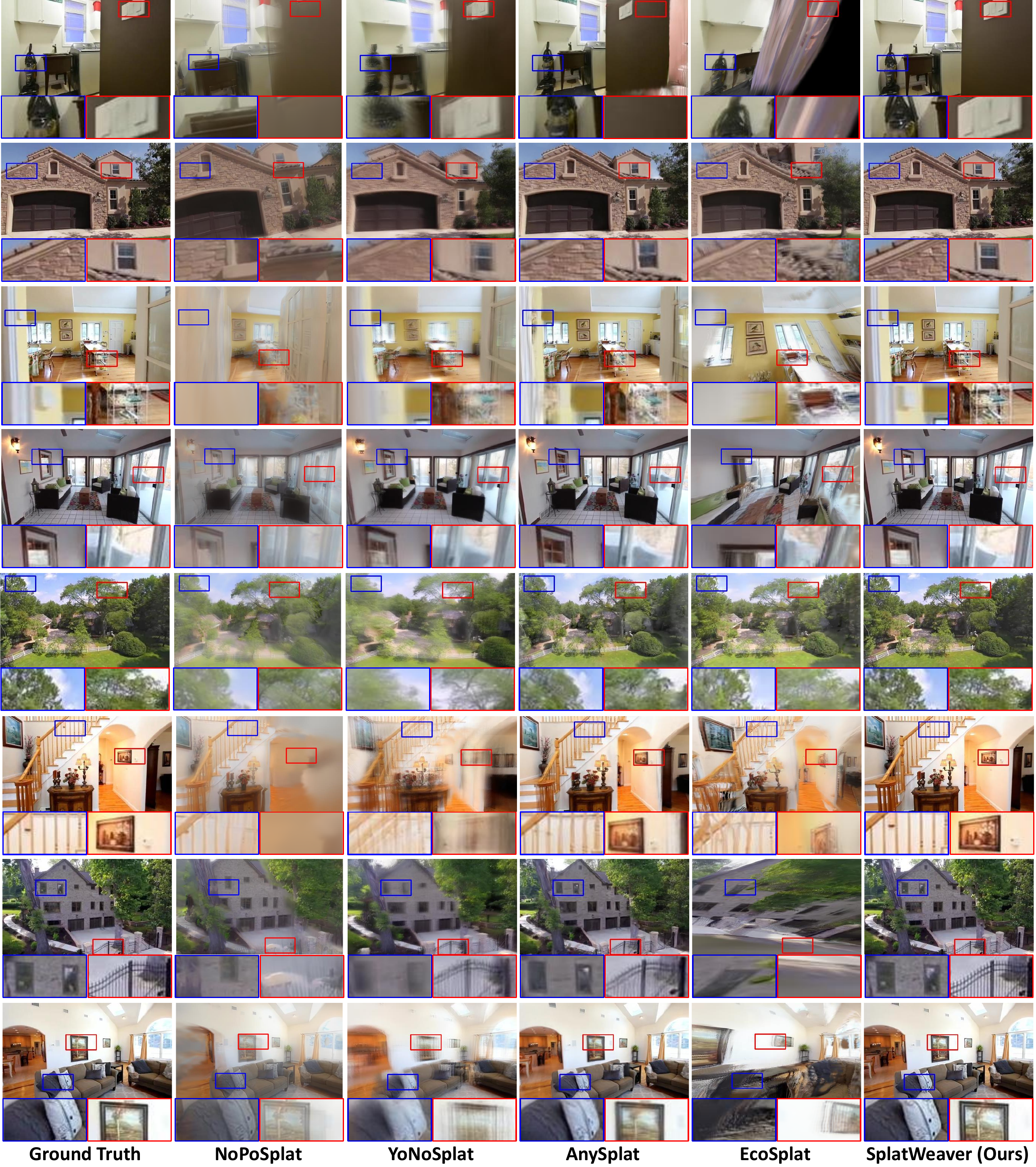}
	\end{center}
	\vspace{-7pt}
	\caption{ Qualitative comparisons on the RealEstate10K \cite{zhou2018stereo}, dataset. From top to bottom, every two rows correspond to rendering results under 4, 8, 16, and 24 view settings, respectively. Our method still generates the most accurate 3D scenes, preserving both photo-realistic texture and geometric-level details. }
	\label{figure_v2}
	\vspace{-10pt}
\end{figure*} 

\begin{figure*}[th!]
	\begin{center}
		\includegraphics[width=.97\linewidth]{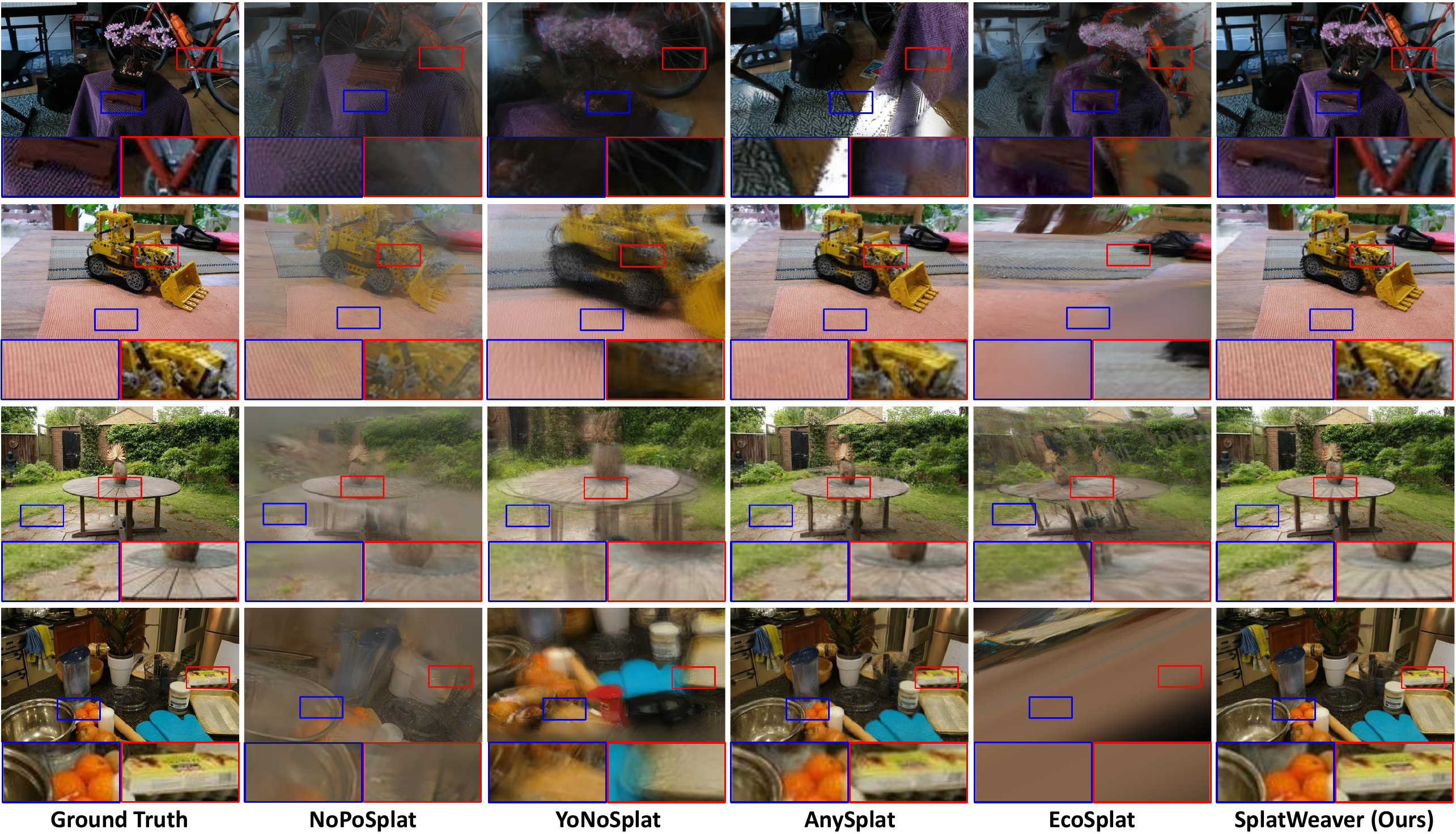}
	\end{center}
	\vspace{-7pt}
	\caption{ Qualitative comparisons on the the Mip-NeRF 360 \cite{barron2022mip} dataset. From top to bottom, each row corresponds to a rendering result under 4, 8, 16, and 24 view settings, respectively. Our method consistently delivers more coherent
		fine structures than other methods in large-scale scenes.}
	\label{figure_v3}
	\vspace{-10pt}
\end{figure*} 
\subsection{Comparison with State-of-the-Art Models}
To rigorously evaluate the effectiveness of SplatWeaver, we conduct a comprehensive comparative analysis against several state-of-the-art baselines. These include pixel-aligned approaches (NoPoSplat \cite{ye2024no}, FLARE \cite{zhang2025flare}, SPFSplat \cite{huang2025no}), voxel-aligned methods (AnySplat \cite{jiang2025anysplat}), pruning-based frameworks (YoNoSplat \cite{ye2026yonosplat}, EcoSplat \cite{park2025ecosplat}), and query-based ones (C3G \cite{an2025c3g}). We also provide the results of AnySplat combined with the offline post-pruning method LightGaussian \cite{fan2024lightgaussian} for reference. Quantitative results are summarized in Tab. \ref{table1}. It is observed that our SplatWeaver delivers remarkable performance gains and outperforms all competitive methods significantly in terms of PSNR, SSIM, and LPIPS. Especially, SplatWeaver achieves a 1.02 dB performance gain over the top-performing AnySplat while utilizing 70\% fewer Gaussian primitives under the 16-view setting. This efficiency stems from our proposed cardinality Gaussian expert routing scheme, which enables adaptive, on-demand allocation. By mitigating redundancy in smooth regions while dedicating more primitives to geometrically complex areas, our method facilitates high-fidelity modeling of 3D scenes. As a result, fewer Gaussians suffice to deliver superior rendering quality. 
While EcoSplat achieves Gaussian reduction through pruning, it incurs a substantial performance drop. Particularly, we observed that its pruning strategy can induce model instability, often leading to failures in scene reconstruction and camera pose estimation. Besides, C3G models 3D scenes through a query-based paradigm; however, its fixed Gaussian budget severely constrains scalability. In other words, regardless of variations in the number of viewpoints, spatial complexity, or scene coverage, the number of predicted Gaussians remains constant, leading to significant under-representation in large-scale scenes.
In contrast, SplatWeaver yields superior rendering quality with economic Gaussians by employing a more physically-grounded and geometry-aware allocation capacity.
Additionally, SplatWeaver eliminates the need for the manual budget scaling, instead autonomously adapting its budget according to scene complexity and coverage. Moreover, we have further compressed the budget during training, yielding an extremely compact version. It can be observed that SplatWeaver$^\dagger$ achieves competitive or even superior performance compared to state-of-the-art methods while using less than 10\% of the Gaussians. This further validates the effectiveness of the proposed allocation mechanism. Fig. \ref{figure_scatter} more clearly demonstrates that our method achieves superior performance with significantly fewer Gaussian primitives.

We also demonstrate visual comparisons in Fig. \ref{figure_v1}, Fig. \ref{figure_v2}, and Fig. \ref{figure_v3}. As suggested, our method produces a more fine-grained and detail-rich rendering by prioritizing primitive allocation in high-complexity regions.
This effectively preserves intricate textures and sharpness,
whereas existing approaches suffer from either detail distortion or unsatisfactory scene estimation failures.

\vspace{-4pt}
\subsection{Results on Dense Novel View Synthesis}
Beyond the standard sparse-view evaluation, we benchmark our method in dense-view scenarios against two distinct paradigms: optimization-based methods, including 3D-GS \cite{kerbl20233d} and Mip-Splatting \cite{yu2024mip}, and representative generalizable frameworks such as Long-LRM \cite{ziwen2025long} and AnySplat \cite{jiang2025anysplat}. As reported in Tab. \ref{table3}, SplatWeaver consistently outperforms both categories across all metrics. While optimization-based approaches often require precise camera poses, meticulous initialization, and prohibitive training time, they remain prone to overfitting, which frequently manifests as artifacts in novel views. In contrast, our framework not only surpasses these baselines but also exceeds the performance of Long-LRM, despite the latter's reliance on known poses. This superiority is primarily attributed to our physically plausible Gaussian primitive allocation, which enables a higher-fidelity and structurally consistent scene representation.
\begin{table}[h]
	\vspace{-3pt}
	\centering
	\footnotesize
	\setlength{\tabcolsep}{7.5pt}
	\caption{Quantitative comparisons of dense novel view synthesis on Mip-NeRF 360 with 64 views.}
	\label{table3}
	\vspace{-4pt}
	\begin{tabular}{l|cccc}
		\toprule[1pt]
		\textbf{Method} & PSNR $\uparrow$ & SSIM $\uparrow$ & LPIPS $\downarrow$&GS$_\times10^3$\\
		\midrule
		3DGS \cite{kerbl20233d} & 22.35 & 0.658 & 0.255&912\\
		Mip-Splatting \cite{yu2024mip} & 22.32 & 0.664 & 0.250 &\textbf{875}\\
		Long-LRM \cite{ziwen2025long} & 22.45 & 0.663 & 0.281 &4237\\
		AnySplat \cite{jiang2025anysplat} & 22.39 & 0.671 & 0.264&5745 \\
		\textbf{SplatWeaver} & \textbf{22.73} & \textbf{0.694} & \textbf{0.245}& 905\\
		\bottomrule[1pt]
	\end{tabular}
	\vspace{-10pt}
\end{table}

\subsection{Results on Camera Pose Estimation}
We further evaluate the performance of our method in camera pose estimation. It is observed from Tab. \ref{table4} that our approach outperforms both VGGT and AnySplat. It is worth noting that while both AnySplat and our method utilize VGGT as the supervision signal, our framework achieves superior pose accuracy. We attribute this gain to our adaptive Gaussian allocation strategy; by reconstructing a sparser yet more representative Gaussian scene, the model can extract more reliable geometric priors, which in turn facilitates more precise camera registration and minimize localization errors.
\begin{table}[h]
	\centering
	\footnotesize
	\setlength{\tabcolsep}{5pt}
	\caption{Camera pose estimation on the RealEstate10K and Co3Dv2 with 10 random frames. }
	\label{table4}
	\vspace{-4pt}
	\begin{tabular}{l|cccc}
		\toprule[1pt]
		\multirow{2}{*}{\textbf{Method}} & \multicolumn{2}{c}{RealEstate10K} & \multicolumn{2}{c}{Co3Dv2} \\
		&AUC@30 $\uparrow$&AUC@10 $\uparrow$&AUC@30 $\uparrow$&AUC@10 $\uparrow$\\
		\midrule
		VGGT \cite{wang2025vggt} & 87.4 &73.2& 76.2& 51.6\\
		AnySplat \cite{jiang2025anysplat} & 87.6&73.1 & 76.8& 52.8\\
		\textbf{SplatWeaver} & \textbf{87.8} &\textbf{73.6}& \textbf{77.9}& \textbf{53.4}\\
		\bottomrule[1pt]
	\end{tabular}
	\vspace{-13pt}
\end{table}

\subsection{Results on Pose-Known Novel View Synthesis}
Another line of work focus on pose-known novel view synthesis, and we further present a quantitative comparison with the feed-forward
novel view synthesis methods \cite{charatan2024pixelsplat,chen2024mvsplat,zhang2025transplat,tang2024hisplat,xu2025depthsplat,zhang2024gs,ziwen2025long,tokengs2026} on RealEstate10K at a $256\times256$ resolution with 2 input views, a setting commonly used in prior works. We adopt DepthSplat \cite{xu2025depthsplat} as the backbone, integrating it with the proposed SplatWeaver architecture and retraining it under the original settings. As evidenced by Tab. \ref{table_re}, our method maintains superior performance even in low-resolution, sparse-view configurations. Notably, we achieve a PSNR improvement of 0.52 dB over the current state-of-the-art, Long-LRM. Furthermore, our approach yields a significantly more compact scene representation, utilizing far fewer Gaussian elements through a more principled and efficient allocation strategy.

\begin{table}[h]
	\vspace{-3pt}	
	\caption{Quantitative comparisons on the RealEstate10K dataset (posed 2 views). * indicates that our model is trained separately with the setting of prior literature \cite{xu2025depthsplat} for fair comparison. }
	\tabcolsep=9pt
	\vspace{-7pt}
	\footnotesize
	\begin{center}
		\begin{tabular}{l|cccc}
			\toprule[1pt]
			\multirow{2}{*}{\textbf{Method}}
			& \multicolumn{4}{c}{\textbf{RealEstate10K}} \\
			& PSNR $\uparrow$ & SSIM $\uparrow$ & LPIPS $\downarrow$ & GS$_\times 10^3$\\
			\midrule
			PixelSplat \cite{charatan2024pixelsplat}& 25.89 & 0.858 & 0.142&131 \\
			MVSplat \cite{chen2024mvsplat}    & 26.39 & 0.869 & 0.128 &131\\
			TranSplat  \cite{zhang2025transplat}  & 26.69 & 0.875 & 0.125&131 \\
			DepthSplat \cite{xu2025depthsplat} & 27.47 &0.889 &0.114&131\\
			GS-LRM \cite{zhang2024gs} &28.10& 0.892& 0.114&131\\
			HiSplat  \cite{tang2024hisplat}               & 27.21 & 0.881 & 0.117&172 \\
            TokenGS \cite{tokengs2026} &28.41&0.903&0.135&262\\
			Long-LRM \cite{ziwen2025long}& 28.54& 0.895& 0.109&117\\
			\textbf{SplatWeaver*}&	\textbf{29.06}&	\textbf{0.899}&	\textbf{0.102}&\textbf{47}\\
			\bottomrule[1pt]
		\end{tabular}
	\end{center}
	\label{table_re}
	\vspace{-10pt}
\end{table} 

\subsection{Efficiency Comparisons}
We also present an efficiency comparison with the existing feed-forward methods in Tab. \ref{table_speed}. It is observed that by implementing a more rational allocation of Gaussian primitives, SplatWeaver achieves the best rendering quality, outperforming all competing methods in terms of PSNR, while simultaneously maintaining a highly compact scene representation. This compactness translates into the lowest storage requirement and the highest rendering speed. These findings confirm that adaptive Gaussian allocation not only enhances rendering fidelity, but also enables a more efficient and scalable 3D representation for generalizable novel view synthesis.
\begin{table}[h]
	\vspace{-3pt}	
	\caption{We report the efficiency metrics
		of the existing methods under the 16-view setting. }
	\tabcolsep=2.8pt
	\vspace{-7pt}
	\footnotesize
	\begin{center}
		\begin{tabular}{c|ccccc}
			\toprule[1pt]
			Method & Latency (s)$\downarrow$ & GS$_\times 10^3$$\downarrow$ & Storage (MB)$\downarrow$ & FPS$\uparrow$ & PSNR$\uparrow$ \\
			\midrule
			NoPoSplat \cite{ye2024no} & 2.7 & 1835 & 119.0 & 191 & 13.88 \\
			AnySplat \cite{jiang2025anysplat}& 1.6 & 1522 & 98.7 & 222 & 19.09 \\
			EcoSplat \cite{park2025ecosplat}& 0.7 & 734 & 47.6 & 275 & 14.64 \\
			SplatWeaver & 1.9 & 451 & 29.2 & 301 & 20.11 \\
			\bottomrule[1pt]
		\end{tabular}
	\end{center}
	\label{table_speed}
	\vspace{-13pt}
\end{table}

\subsection{Empirical Analyses}\label{abla}
In Tab. \ref{table_pc}, we conduct ablation experiments on the dedicated components introduced in SplatWeaver. The
effectiveness of each proposed component is evaluated by systematically integrating them into the model, revealing their individual contributions. Detailed analyses are provided below.

\begin{table}[h]
	\centering
	\footnotesize
	\setlength{\tabcolsep}{6.5pt}
	\caption{Ablation of the basic model components.}
	\label{table_pc}
	\vspace{-3pt}
	\begin{tabular}{lccc}
		\toprule[1pt]
		\multirow{2}{*}{\textbf{Variant}} & \multicolumn{3}{c}{\textbf{Dl3DV} (16 Views)} \\
		& PSNR $\uparrow$ & SSIM $\uparrow$ & LPIPS $\downarrow$ \\
		\midrule
		Baseline (naive pruning) & 17.56 & 0.488 & 0.402 \\
		+ Cardinality Gaussian Expert & 19.19 & 0.552 & 0.299 \\
		+ Frequency Prior Guidance & 19.77 & 0.591 & 0.273 \\
		+ Neighbor-Conditioned Prediction & 20.11 & 0.607 & 0.260 \\
		\bottomrule[1pt]
	\end{tabular}
	\vspace{-3pt}
\end{table}

\begin{figure*}[t]
	\begin{center}
		\includegraphics[width=\linewidth]{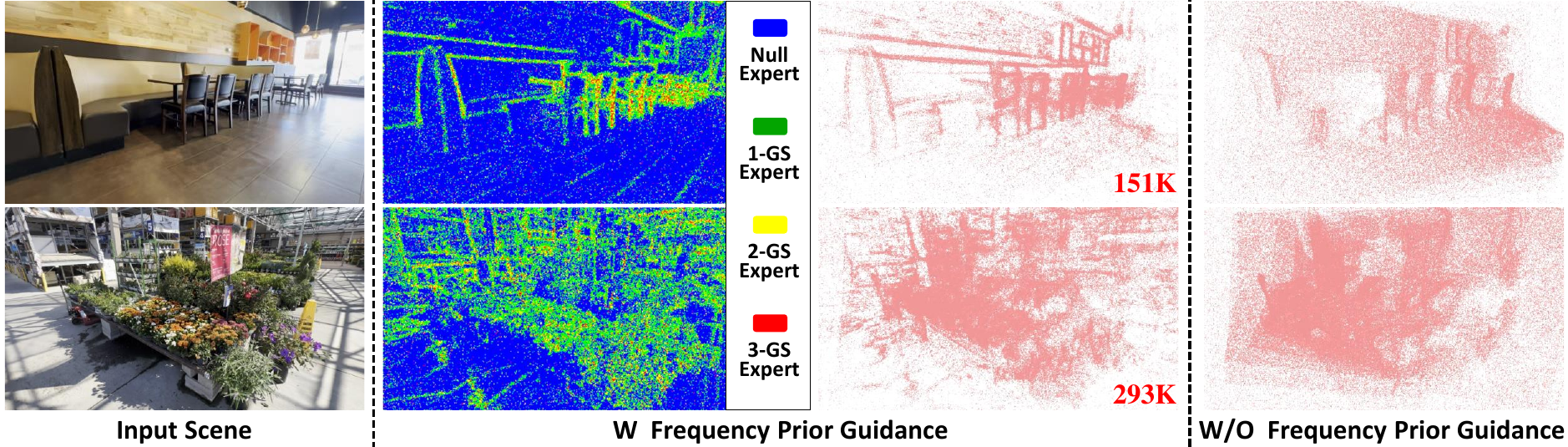}
	\end{center}
	\vspace{-7pt}
	\caption{ Visualization of the cardinality Gaussian expert routing and the resulting Gaussian distribution with or without the frequency prior guidance (network module and regularization loss).}
	\label{figure5}
	\vspace{-10pt}
\end{figure*} 

\noindent\textbf{Effect of cardinality Gaussian expert routing.}
To enable adaptive feed-forward Gaussian allocation, we propose the cardinality Gaussian expert routing scheme. As evidenced in Tab. \ref{table_pc}, this scheme yields a substantial gain of 1.63 dB in PSNR, underscoring the pivotal role of adaptive Gaussian allocation in high-quality forward 3D reconstruction.

In Fig. \ref{figure5}, we visualize the expert routing results for different scenes as well as the distribution of Gaussian primitives. It can be observed that our method adaptively routes experts based on the complexity of the scene, adhering to the principle of “\emph{dense where complex, sparse where smooth}” to achieve a more physically reasonable distribution.
In addition, it also reveals that our method can adaptively modify the budget based on the complexity of the scene, allocating more Gaussian primitives to scenes with complex textures and geometry, and deploying fewer Gaussian primitives for simpler scenes.

\begin{table}[h]
	\centering
	\footnotesize
	\setlength{\tabcolsep}{13.5pt}
	\caption{Ablation of the number of experts.}
	\label{table_en}
	\vspace{-3pt}
	\begin{tabular}{c|ccc}
		\toprule[1pt]
		\multirow{2}{*}{\textbf{Experts Number}} & \multicolumn{3}{c}{\textbf{Dl3DV} (16 Views)} \\
		& PSNR $\uparrow$ & SSIM $\uparrow$ & LPIPS $\downarrow$ \\
		\midrule
		2 & 19.23 & 0.562 & 0.282 \\
		3 & 19.57 & 0.581 & 0.272 \\
		4 & 20.11 & 0.607 & 0.260 \\
		5 & 20.05 & 0.607 & 0.265 \\
		\bottomrule[1pt]
	\end{tabular}
	\vspace{-3pt}
\end{table}

In addition, we investigate the impact of the number of experts in Tab. \ref{table_en}. As observed, an insufficient number of experts restricts the model's allocation capacity, leading to sub-optimal performance. Nevertheless, even with only two experts, our routing paradigm still surpasses existing opacity-based or score-based pruning methods. This advantage stems from the fact that those methods rely on Gaussian importance learning and passive pruning, whereas our method can allocate Gaussians more flexibly and adaptively on demand. This flexible allocation capability results in more efficient 3D representation.
Additionally, while increasing the number of experts initially yields steady improvements, a slight performance degradation occurs when the count reaches 5. We attribute this to the fact that four experts already provide sufficient capacity to capture variations in scene complexity, whereas introducing additional experts increases optimization difficulty within a higher-dimensional allocation space.

\noindent\textbf{Effect of frequency prior guidance.} As illustrated in Tab. \ref{table_pc}, the proposed frequency prior guidance strategy (guidance module and regularization) can effectively guide the model to allocate Gaussians according to regional complexity, delivering a significant performance gain of 0.58 dB. Furthermore, Fig. \ref{figure5} demonstrates that the proposed scheme facilitates a more rational and physically plausible allocation, whereas eliminating it prevents the model from capturing the intrinsic mapping between scene complexity and Gaussian density, resulting in a suboptimal allocation.

\noindent\textbf{Effect of neighbor-conditioned prediction.} Instead of directly regressing the complete set of Gaussian parameters, our framework decomposes the estimation process into two stages. The Gaussian expert is responsible for predicting only the spatial locations along with their associated latent features, while the remaining attributes are inferred through the aggregation of features from spatial neighbors, leading to a more expressive and refined representation. As shown in Tab. \ref{table_pc}, this spatial context modeling brings a 0.24 dB improvement, substantiating its effectiveness in capturing local geometric correlations.

Additionally, we evaluate the influence of the number of neighbors, $k$, on our neighbor-conditioned Gaussian parameter prediction. As evidenced in Tab. \ref{table_neighbor}, our method is relatively insensitive to the choice of $k$. Performance peaks at $k=8$, beyond which we observe diminishing returns in accuracy. Consequently, we set $k=8$ as the default to balance reconstruction quality and computational efficiency.
\begin{table}[h]
	\vspace{-3pt}	
	\caption{Ablation of the number of neighboring Gaussians. }
	\tabcolsep=12.6pt
	\vspace{-3pt}
	\footnotesize
	\vspace{-4pt}
	\begin{center}
		\begin{tabular}{c|cccc}
			\toprule[1pt]
			\multirow{2}{*}{\textbf{$k$}}& \multicolumn{4}{c}{\textbf{Dl3DV} (16 Views)}\\
			&PSNR $\uparrow$ & SSIM $\uparrow$ & LPIPS $\downarrow$& latency (s)\\
			\midrule
			4 &19.88&0.597&0.268&0.09\\
			6 &19.95&0.604&0.262&0.10\\
			8 &20.11&0.607&0.260&0.11\\
			10 &20.12&0.607&0.260&0.12\\
			\bottomrule[1pt]
		\end{tabular}
	\end{center}
	\label{table_neighbor}
	\vspace{-10pt}
\end{table}

\noindent\textbf{Ablation Study of $\rho_3\%$, $\rho_2\%$, and $\rho_1\%$.} As detailed in Tab. \ref{table_rho}, the parameters $\rho_1, \rho_2$, and $\rho_3$ regulate the distribution of different expert types. We observe that excessively high values for $\rho_3$ and $\rho_2$ can disrupt the allocation balance in smoothing regions, leading to insufficient representation and suboptimal performance. However, within a reasonable operational range, our framework exhibits remarkable robustness to these proportions, as this constraint is only deployed during the initial half of training to provide a reasonable guidance, whereas in the latter half, the model explores the optimal complexity-aware allocation strategy autonomously.
\begin{table}[h]
	\vspace{-3pt}    
	\caption{Ablation study of $\rho_3\%$, $\rho_2\%$, and $\rho_1\%$ with fixed Gaussian budget. }
	\label{table_rho}
	\vspace{-7pt}
		\tabcolsep=10pt
	\footnotesize
	\begin{center}
		\begin{tabular}{ccc|ccc}
			\toprule[1pt]
			\multirow{2}{*}{\textbf{$\rho_3\%$}} & \multirow{2}{*}{\textbf{$\rho_2\%$}} & \multirow{2}{*}{\textbf{$\rho_1\%$}} & \multicolumn{3}{c}{\textbf{Dl3DV} (16 Views)} \\
			& & & PSNR $\uparrow$ & SSIM $\uparrow$ & LPIPS $\downarrow$ \\
			\midrule
			0.10 & 0.00 & 0.00 & 19.57 & 0.576 & 0.302 \\
			0.05 & 0.05 & 0.05 & 19.87 & 0.593 & 0.287 \\
			0.01 & 0.01 & 0.25 & 20.06 & 0.611 & 0.266 \\
			0.02 & 0.02 & 0.20 & 20.11 & 0.607 & 0.260 \\
			\bottomrule[1pt]
		\end{tabular}
	\end{center}
	\vspace{-10pt} 
\end{table}

\noindent\textbf{Ablation Study of $\epsilon$.} 
The parameter $\epsilon$ governs the total budget of Gaussian primitives. Our empirical analysis reveals considerable redundancy within scene representations; although increasing the primitive count allows for a more exhaustive modeling of fine details, the resulting marginal gains in performance are not cost-effective. Furthermore, an excessive number of Gaussians imposes an unnecessary computational footprint on both storage and rendering efficiency. As demonstrated in Tab. \ref{table_epsilon}, the optimal equilibrium between reconstruction fidelity and efficiency is achieved when the total budget is set to 0.3 times the pixel count.
\begin{table}[h]
	\vspace{-3pt}	
	\caption{Ablation of Gaussian budget control factor $\epsilon$. }
	\tabcolsep=12.8pt
	\vspace{-7pt}
	\footnotesize
	\begin{center}
		\begin{tabular}{c|cccc}
			\toprule[1pt]
			\multirow{2}{*}{\textbf{$\epsilon$}}& \multicolumn{4}{c}{\textbf{Dl3DV} (16 Views)}\\
			&PSNR $\uparrow$ & SSIM $\uparrow$ & LPIPS $\downarrow$& GS$_\times 10^3$\\
			\midrule
			0.1 &19.52&0.587&0.279&153\\
			0.3 &20.11&0.607&0.260&451\\
			0.5 &20.17&0.617&0.257&868\\
			1.0 &20.47&0.629&0.241&1744\\
			\bottomrule[1pt]
		\end{tabular}
	\end{center}
	\label{table_epsilon}
	\vspace{-10pt}
\end{table}

\begin{figure}[h]
	\begin{center}
		\includegraphics[width=.95\linewidth]{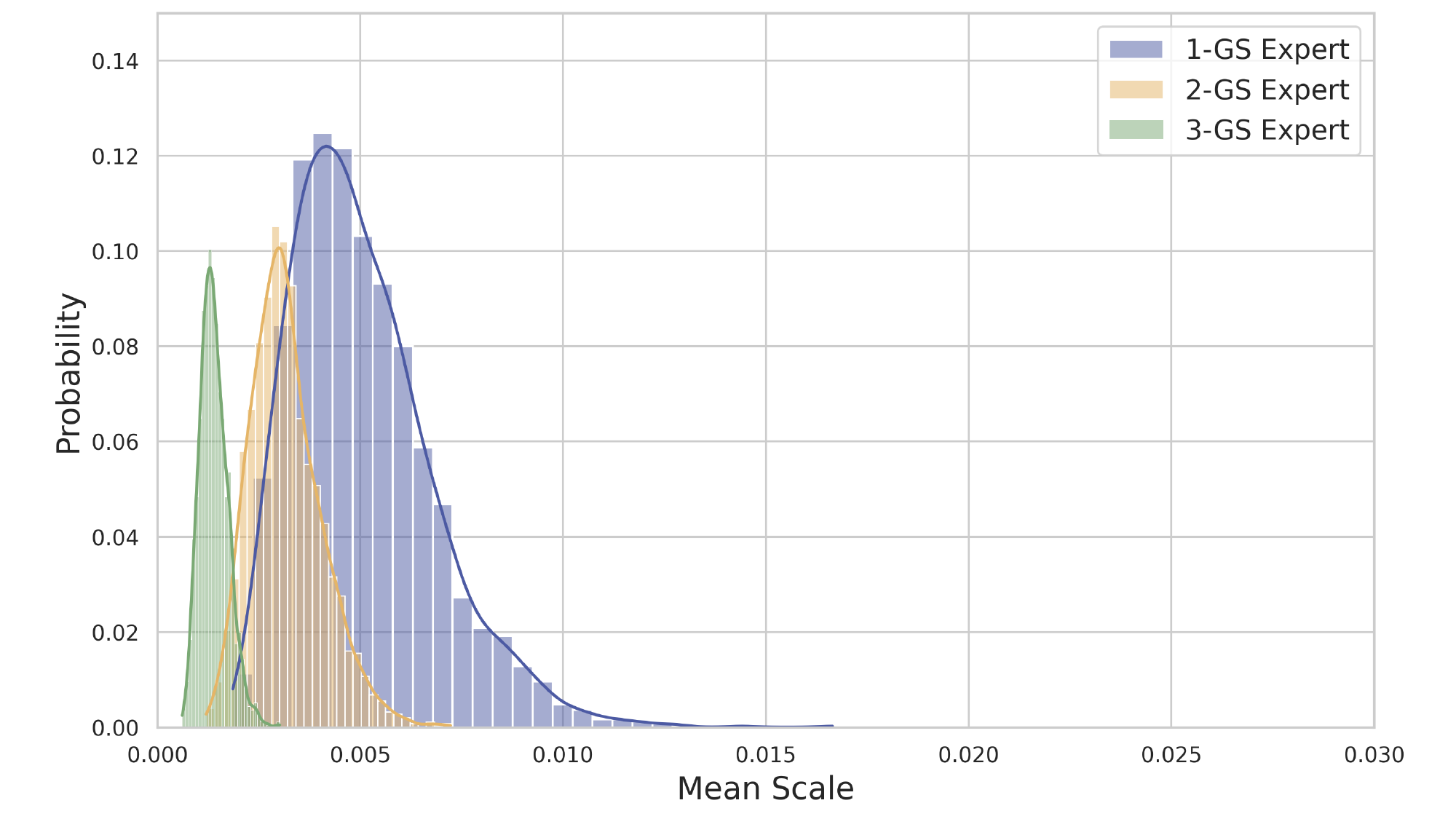}
	\end{center}
	\vspace{-7pt}
	\caption{ Visualization of Gaussian scales predicted across different experts.}
	\label{figure6}
	\vspace{-6pt}
\end{figure}

\noindent\textbf{Visualization of Gaussian Scales Predicted Across Different Experts.} In Fig. \ref{figure6}, we visualize the distribution of Gaussian scales predicted by different cardinality Gaussian experts. It is observed that low-cardinality experts predominantly specialize in smooth regions, generating large-scale Gaussian primitives to efficiently cover homogeneous areas. Conversely, high-cardinality experts focus on intricate structures within complex regions, producing fine-grained, small-scale primitives to capture high-frequency geometric details. This distinct specialization aligns with geometric intuition and underscores the physical plausibility of our adaptive allocation framework.

\begin{figure}[h]
	\begin{center}
		\includegraphics[width=\linewidth]{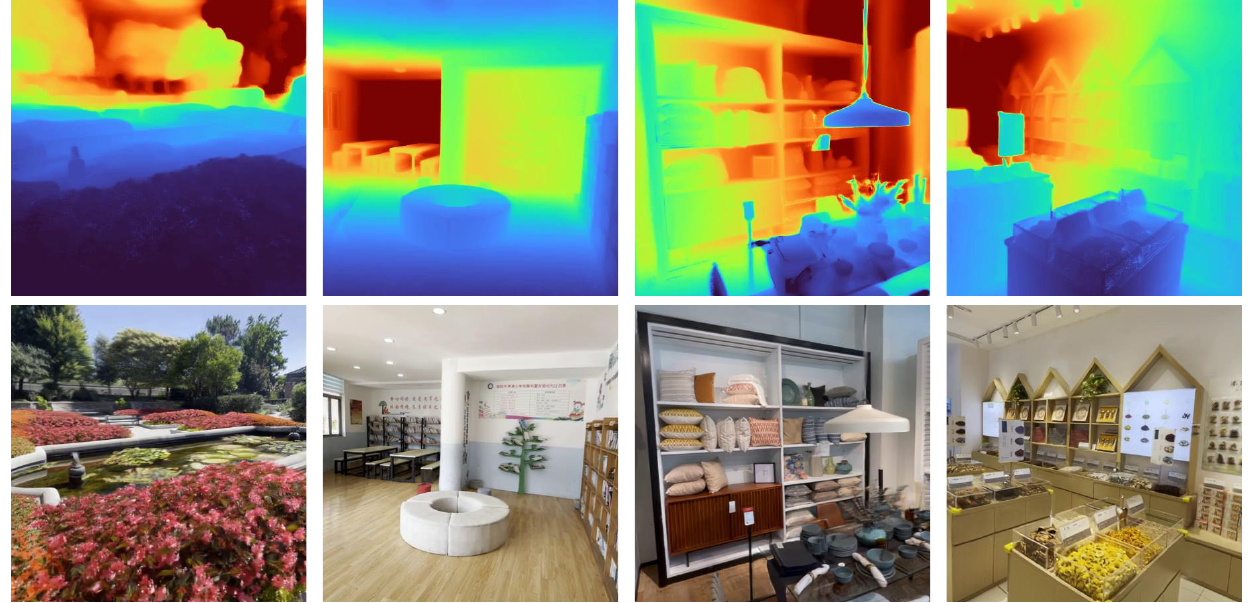}
	\end{center}
	\vspace{-7pt}
	\caption{ Visualization of scene geometry and novel view synthesis.}
	\label{figure7}
	\vspace{-16pt}
\end{figure}

\noindent\textbf{Visualization of Scene Geometry.} While our approach forgoes the conventional per-pixel Gaussian modeling paradigm commonly adopted in existing methods, it nevertheless achieves superior geometric reconstruction through a physically plausible, non-uniform primitive distribution. As illustrated in Fig. \ref{figure7}, SplatWeaver not only delivers high-fidelity novel view synthesis but also generates detailed and accurate depth maps. This capability underscores the structural fidelity of our adaptive allocation mechanism and further validates the effectiveness of our sparse-yet-precise scene representation.

\section{Concluding Remarks}\label{5}
In this work, we propose SplatWeaver, an innovative framework that enables efficient and adaptive allocation of Gaussian primitives in a feed-forward manner. In contrast to existing methods that typically predict uniform per-pixel or per-voxel Gaussian primitives, which often suffer from redundancy in simple regions and deficiency in complex ones, our approach elegantly tackles these challenges through a dedicated cardinality Gaussian expert routing scheme. This routing paradigm allows the model to not only eliminate redundancy but also concentrate Gaussians on detail-rich areas, resulting in a more expressive 3D scene representation.
Extensive experiments on various novel view synthesis benchmarks manifest the effectiveness, superiority, and efficiency of our method.
We expect this work to provide insights into more effective generalizable novel view synthesis and steer future research on this Gordian knot.

\bibliographystyle{IEEEtran}
\bibliography{egbib}

\vfill

\end{document}